\documentclass[review]{elsarticle}

\usepackage[colorlinks,linkcolor=blue,citecolor=teal]{hyperref}  
\usepackage{float}
\usepackage{bm}
\usepackage{amsmath}
\numberwithin{equation}{section}
\usepackage{amsfonts}
\usepackage{amsthm}
\usepackage{amssymb}

\usepackage{tikz-cd}
\theoremstyle{definition}

\usepackage{xcolor}

\definecolor{revisionblue}{RGB}{0,90,180}

\newenvironment{revision}{\color{revisionblue}}{}
\begin{document}
\begin{frontmatter}

\title{Scalable Perturbation Learning for Online Self-Supervised Learning in Echo State Networks}
\author[utokyo]{Taiki Yamada\corref{cor1}}
\ead{yamada-taiki@g.ecc.u-tokyo.ac.jp}
\author[utokyo]{Kantaro Fujiwara}
\ead{kantaro@g.ecc.u-tokyo.ac.jp}

\cortext[cor1]{Corresponding author}

\affiliation[utokyo]{
  organization={Graduate School of Information Science and Technology, The University of Tokyo},
  addressline={7-3-1 Hongo},
  city={Bunkyo-ku},
  postcode={113-8656},
  state={Tokyo},
  country={Japan}
}

\begin{abstract}
Intelligent systems should not only solve tasks but also adapt under real-world constraints. 
Autonomous adaptation via self-supervised learning, sequential adaptation via online learning, and memory-efficient implementation via perturbation-based learning are important requirements for such systems.
However, these requirements are generally in tension for high-dimensional systems.
In this study, we focus on echo state networks (ESNs), where this tension naturally arises in large reservoirs. 
We propose a perturbation-based learning rule for online self-supervised learning in ESNs. 
The proposed rule is derived from an orthogonal decomposition of the self-supervised learning loss, which separates an input-dependent component from a redundant component determined by the fixed ESN parameters. 
Perturbing only the input-dependent component reduces the effective perturbation dimension from the reservoir dimension to the input dimension.
Thus, the proposed method preserves self-supervised adaptation, online learning, and scalar-feedback perturbation learning, while avoiding reservoir-size-dependent variance growth. 
This suggests a design principle for scalable and hardware-compatible learning: online learning should be restricted to the dynamically necessary low-dimensional component of the objective.
\end{abstract}

\end{frontmatter}


\section{Introduction}

Building intelligent systems that can continuously adapt to complex real-world dynamics is an important goal in machine learning \citep{parisi_continual_2019} and neuromorphic engineering \citep{mead_neuromorphic_1990}.
Such systems must process temporal information under limited computational and memory resources \citep{maass_real-time_2002, indiveri_memory_2015}.
Reservoir computing addresses this challenge by using a recurrent dynamical system as a temporal feature extractor and reducing the trainable part of the model \citep{lukosevicius_reservoir_2009, tanaka_recent_2019}.
Echo state networks (ESNs) \citep{jaeger2001echo} are a representative model of reservoir computing, in which the reservoir is a fixed recurrent neural network.
Realizing such continuously adapting systems with ESNs requires satisfying three conditions.
First, learning should be online and self-supervised, so that the system can adapt without external target labels in nonstationary environments \citep{parisi_continual_2019}.
Second, the reservoir dimension should be scalable, because large reservoirs are often used to represent high-dimensional dynamics and long temporal dependencies \citep{jaeger_adaptive_2002, dambre_information_2012}.
Third, this scaling should not rely on large auxiliary memory or complex error routing \citep{indiveri_memory_2015, lillicrap_backpropagation_2020, lin_-device_2022}.
This requirement is especially important for hardware implementations that exploit reservoir dynamics directly, rather than relying on general-purpose computation and full memory access.

\begin{figure}[H]
\begin{center}
\includegraphics[width=\linewidth]{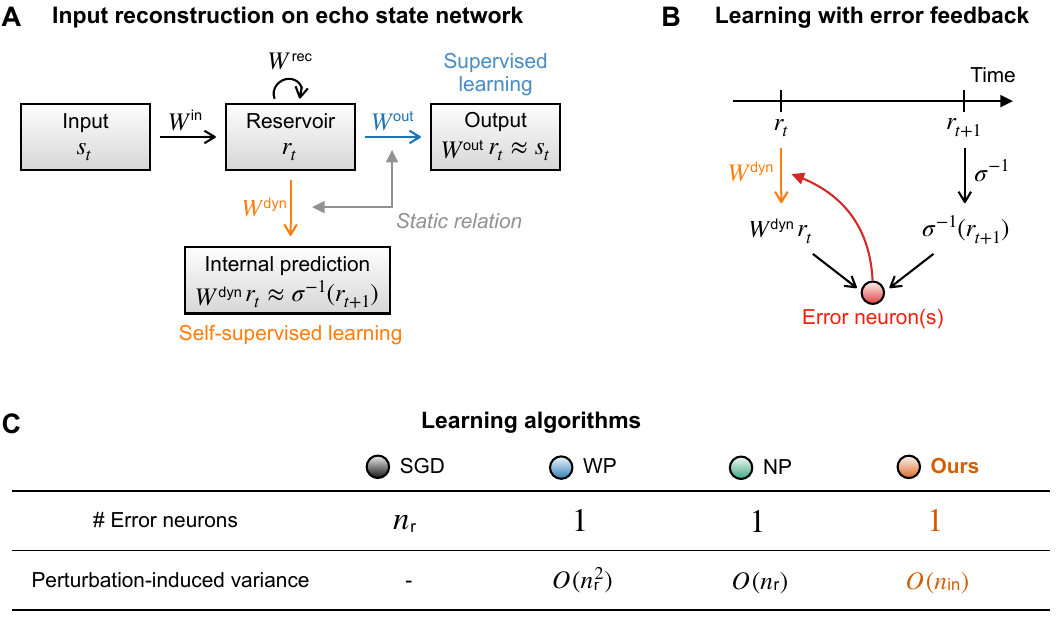}
\caption{Overview of the proposed learning rule.
(A) The input reconstruction task on an echo state network, viewed in two ways. A supervised readout $W^{\mathrm{out}}$ maps the reservoir state $r_t$ back to the input, while the self-supervised formulation instead predicts the inverse-activated next reservoir state from the current one, $W^{\mathrm{dyn}}r_t \approx \sigma^{-1}(r_{t+1})$, using a target available internally from the reservoir dynamics; the two are linked by the static transformation $\mathcal{P}$.
(B) In the self-supervised setting, each reconstruction unit receives its own error signal, so learning with per-unit error feedback uses $O(n_{\mathrm{r}})$ error neurons.
(C) The learning algorithms are compared in terms of the number of error neurons and the perturbation-induced variance. SGD uses $n_{\mathrm{r}}$ error neurons and, not being perturbation-based, incurs no perturbation-induced variance ($-$); weight (WP) and node (NP) perturbation use a single global scalar signal instead, with variance scaling as $O(n_{\mathrm{r}}^2)$ and $O(n_{\mathrm{r}})$, respectively; the proposed rule (Ours) keeps the single scalar signal but attains $O(n_{\mathrm{in}})$ variance.}
\label{fig:overview}
\end{center}
\end{figure}

However, these requirements are in tension.
To explain this point, we denote the input, reservoir, and output dimensions by \(n_{\mathrm{in}}\), \(n_{\mathrm{r}}\), and \(n_{\mathrm{out}}\), respectively.
Although ESNs can be applied to various temporal tasks by supervised readout learning \citep{jaeger2001echo}, only specific tasks admit a self-supervised formulation.
In this paper, we consider the task studied in our prior work \citep{yamada_unsupervised_2026}, where the system learns to recover the external input from the reservoir dynamics without external teaching signals (Figure~\ref{fig:overview}A).
For this task, \(n_{\mathrm{out}}=n_{\mathrm{in}}\).
Our prior work reformulated this problem not as training the usual \(n_{\mathrm{in}} \times n_{\mathrm{r}}\) output map, but as training an \(n_{\mathrm{r}} \times n_{\mathrm{r}}\) map that reconstructs the reservoir state itself.
As a consequence of this self-supervised formulation, the reconstruction target and the reconstruction error both lie in the \(n_{\mathrm{r}}\)-dimensional reservoir space.

Batch least squares \citep{Legendre1805, Gauss1809} is not a suitable learning algorithm for this setting, because it requires storing data and solving a global inverse problem after data collection, and so is not online \citep{jaeger2001echo, jaeger_harnessing_2004, lukosevicius_reservoir_2009}.
Recursive least squares \citep{plackett_theorems_1950} makes the update online, but it maintains a precision matrix of size \(O(n_{\mathrm{r}}^2)\) in the self-supervised learning setting, which incurs a quadratic memory cost \citep{sussillo_generating_2009, lukosevicius_reservoir_2009}.
Stochastic gradient descent \citep{robbins_stochastic_1951, bottou_large-scale_2010} can be viewed as a sample-wise online alternative that avoids this quadratic state.
However, it still requires an \(n_{\mathrm{r}}\)-dimensional error signal at each time step in the self-supervised learning setting.
Thus, even when the memory cost is reduced, learning still requires \(O(n_{\mathrm{r}})\) error buffers and feedback channels to provide distinct error signals to the reconstruction units (Figure~\ref{fig:overview}B).

Perturbation-based learning \citep{kiefer_stochastic_1952} provides a possible way to avoid such unit-wise error feedback, but its scalar feedback introduces a variance scaling problem \citep{werfel_learning_2005, ren2023scaling}.
In weight perturbation \citep{jabri_weight_1991} and node perturbation \citep{flower_summed_1992, fiete_gradient_2006}, parameter updates are estimated from random perturbations and a scalar change in the objective function.
Thus, the \(n_{\mathrm{r}}\)-dimensional reconstruction error is compressed into a single global scalar signal.
This removes the need for multi-channel error routing and replaces it with a scalar feedback signal that can be broadcast globally \citep{jabri_weight_1991, flower_summed_1992, fiete_gradient_2006}.
However, the compression of an error vector into a scalar signal by perturbation-based methods introduces additional variance into the gradient estimate, which scales with network size \citep{werfel_learning_2005, ren2023scaling}.
In the present $n_{\mathrm{r}} \times n_{\mathrm{r}}$ learning problem, this additional variance scales as $O(n_{\mathrm{r}}^2)$ for weight perturbation and $O(n_{\mathrm{r}})$ for node perturbation.
Thus, as $n_{\mathrm{r}}$ increases, the signal-to-noise ratio of the gradient estimate decreases, making perturbation-based learning inefficient for high-dimensional reservoirs.

In this paper, we address this variance scaling problem by exploiting the structure of online self-supervised learning in ESNs.
We show that the self-supervised loss function admits an orthogonal decomposition into two components.
One component can be handled by a fixed transformation that is independent of the online input sequence, whereas the remaining component is the only part that has to be learned from online data.
This decomposition reveals that the online error representation and perturbation source required for the input-dependent component are both $n_{\mathrm{in}}$-dimensional rather than $n_{\mathrm{r}}$-dimensional.
Based on this observation, we apply perturbation-based learning only to the input-dependent component and recover the full self-supervised solution through the fixed transformation.
As a result, the perturbation-induced variance scaling is reduced from the $O(n_{\mathrm{r}}^2)$ of conventional weight perturbation or the $O(n_{\mathrm{r}})$ of node perturbation to $O(n_{\mathrm{in}})$ (Figure~\ref{fig:overview}C).
Thus, when $n_{\mathrm{in}} \ll n_{\mathrm{r}}$, the proposed learning rule remains scalable even for large reservoirs.
It preserves the scalar global feedback mechanism of perturbation-based learning while avoiding reservoir-dimensional error routing and reducing the variance cost of online self-supervised learning.
This makes the proposed approach suitable for large ESNs and for hardware settings where memory access and feedback routing are limited.

The remainder of this paper is organized as follows. 
Section~\ref{sec:theory} presents the theoretical formulation of online self-supervised learning in ESNs.
Section~\ref{sec:learning-alg} reviews general online learning rules, namely gradient descent, stochastic gradient descent, weight perturbation, and node perturbation.
Section~\ref{sec:sSL-ESN} reviews the self-supervised formulation of ESNs introduced in our prior work, on which this study is based.
Section~\ref{sec:proposed} presents our main contribution, namely the orthogonal loss decomposition and the resulting perturbation-based learning rule.
Section~\ref{sec:numerical} verifies the theoretically predicted scaling behavior through numerical experiments and compares the proposed rule with existing learning rules.
Section~\ref{sec:discussion} discusses the implications and limitations of the proposed rule.
Section~\ref{sec:conclusion} concludes the paper.

\newcommand{\R}{\mathbb{R}}
\newcommand{\Win}{W^{\mathrm{in}}}
\newcommand{\Wr}{W^{\mathrm{rec}}}
\newcommand{\Wout }{W^{\mathrm{out}}}
\newcommand{\Wdyn}{W^{\mathrm{dyn}}}

\section{Theoretical Formulation}
\label{sec:theory}
Throughout this paper, $\langle \cdot \rangle_u$ denotes the expectation with respect to the distribution of the random variable $u$.
For symmetric positive semidefinite matrices $G$ and $H$, we define the induced seminorms
\begin{equation}
    \begin{aligned}
    \|x\|_G^2 &:= x^\top G x, \\
    \|X\|_{G,H}^2 &:= \mathrm{tr}\!\left(G X H X^\top\right).
    \end{aligned}
\end{equation}
These become norms when the corresponding matrices are positive definite, and we write $\|x\| := \|x\|_I$ and $\|X\| := \|X\|_{I,I}$ for the standard Euclidean and Frobenius norms.

\subsection{Learning Algorithms}
\label{sec:learning-alg}
We recall a general formalization of online learning rules. 
As a reference, gradient descent \citep{Cauchy1847} minimizes the loss using the full-batch gradient; its online counterparts include stochastic gradient descent \citep{robbins_stochastic_1951}, weight perturbation \citep{jabri_weight_1991}, and node perturbation \citep{flower_summed_1992, fiete_gradient_2006}.
Consider a system with a parameter matrix $W$ to be trained and an input $z$.
Let $\ell(W;z)\in\R$ denote the instantaneous loss for an input $z$, and define the expected objective by
\begin{equation}
    L(W)=\langle \ell(W;z)\rangle_z.
\end{equation}
We consider the problem of finding $W$ that minimizes the expected objective $L(W)$.
The update rule of gradient descent (GD) \citep{Cauchy1847} for the objective function $L$ is defined as
\begin{alignat}{1}
    \label{eq:GD}
    \Delta^{\mathrm{GD}} W
    = -\eta\frac{\partial L(W)}{\partial W}
    = -\eta \left\langle \frac{\partial \ell(W;z)}{\partial W} \right\rangle_z,
\end{alignat}
where \(\eta>0\) is the learning rate.

If the true distribution of $z$ is unknown, the expectation $\langle\cdot\rangle_z$ in Eq.~(\ref{eq:GD}) cannot be evaluated directly.
Stochastic gradient descent (SGD) \citep{robbins_stochastic_1951} approximates GD by replacing the expectation with the instantaneous value at an observed sample $z$.
In particular, given an input sample $z$, SGD for the instantaneous loss $\ell(W;z)$ is defined by
\begin{alignat}{1}
    \label{eq:SGD}
    \Delta^{\mathrm{SGD}}W
    =
    -\eta \frac{\partial \ell(W;z)}{\partial W}.
\end{alignat}
The SGD update in Eq. (\ref{eq:SGD}) is an unbiased estimator of the GD update:
\begin{alignat}{1}
\left\langle \Delta^{\mathrm{SGD}} W \right\rangle_z
= \Delta^{\mathrm{GD}} W.
\end{alignat}
Since the estimator is unbiased, its deviation from GD is due to stochastic fluctuations around its mean.
We quantify this fluctuation by the estimation variance, defined for a matrix-valued update as the expected squared Frobenius norm of its deviation from the mean:
\begin{equation}
    \begin{aligned}
    \mathrm{Var}_z\!\left[\Delta^{\mathrm{SGD}} W\right]
    &:=
    \left\langle \left\| \Delta^{\mathrm{SGD}} W
    - \left\langle \Delta^{\mathrm{SGD}} W \right\rangle_z \right\|^2 \right\rangle_z \\
    &=
    \left\langle \left\| \Delta^{\mathrm{SGD}} W
    - \Delta^{\mathrm{GD}} W \right\|^2 \right\rangle_z \\
    &=
    \eta^2
    \left\langle \left\|
    \frac{\partial \ell(W;z)}{\partial W}
    - \frac{\partial L(W)}{\partial W}
    \right\|^2 \right\rangle_z .
    \end{aligned}
\end{equation}

A smaller update variance implies that the stochastic update more closely follows the GD update, whereas a large update variance can destabilize the learning dynamics.
Thus, for sufficiently small $\eta>0$, the fluctuations inherited from the randomness of $z$ are averaged out over successive updates, and SGD can provide a stable learning algorithm for minimizing the objective function under appropriate regularity conditions.

Weight perturbation and node perturbation are zeroth-order methods that avoid explicit calculation of derivatives by using numerical differentiation.
Specifically, weight perturbation (WP) \citep{jabri_weight_1991} approximates the derivative $\frac{\partial \ell(W;z)}{\partial W}$ in SGD as follows\footnote{
While zeroth-order updates are sometimes defined with a forward difference
$(\ell(W+\alpha\Xi;z)-\ell(W;z))/\alpha$,
we adopt the central (antithetic) difference
$(\ell(W+\alpha\Xi;z)-\ell(W-\alpha\Xi;z))/(2\alpha)$.
For the quadratic losses considered in this paper, the central difference equals the exact directional derivative for any $\alpha$, so the update is independent of the perturbation scale and the variance formulas below hold exactly; a forward difference would instead retain an $O(\alpha)$ curvature term whose variance grows with an additional factor of the perturbed dimension.
}:
\begin{alignat}{1}
    \label{eq:WP}
    \Delta^{\mathrm{WP}} W
    =
    -\eta
    \frac{\ell(W+\alpha\Xi;z)-\ell(W-\alpha\Xi;z)}{2\alpha}\Xi,
\end{alignat}
where $\Xi$ is a perturbation matrix of the same size as $W$, and $\alpha>0$ is the scale parameter of the perturbation.

We assume that its entries are independent, zero-mean, unit-variance random variables with a common bounded fourth moment:
\begin{equation}
    \left\langle \Xi_{ij} \right\rangle_\Xi = 0,
    \qquad
    \left\langle \Xi_{ij}\Xi_{kl}\right\rangle_\Xi = \delta_{ik}\delta_{jl},
    \qquad
    \left\langle \Xi_{ij}^4 \right\rangle_\Xi = \mu_{4,\Xi}<\infty .
\end{equation}
For sufficiently small $\alpha$, WP approximates the SGD update in expectation over the perturbation:
\begin{equation}
    \label{eq:WP-SGD}
    \begin{aligned}
    \left\langle \Delta^{\mathrm{WP}} W \right\rangle_{\Xi}
    &=\left\langle
    -\eta \frac{\ell(W+\alpha\Xi;z)-\ell(W-\alpha\Xi;z)}{2\alpha}\Xi
    \right\rangle_{\Xi}\\
    &\approx
    -\eta
    \left\langle
    \left( \sum_{ij} \frac{\partial \ell(W;z)}{\partial W_{ij}} \Xi_{ij} \right) \Xi
    \right\rangle_{\Xi}\\
    &= -\eta \frac{\partial \ell(W;z)}{\partial W}\\
    &= \Delta^{\mathrm{SGD}} W.
    \end{aligned}
\end{equation}

The cost of this derivative-free approximation is the random deviation from SGD induced by $\Xi$ for each fixed $z$, which is quantified by the corresponding estimation variance:
\begin{equation}
    \begin{aligned}
    \mathrm{Var}_{\Xi}\!\left[\Delta^{\mathrm{WP}} W \mid z\right]
    &:=
    \left\langle \left\| \Delta^{\mathrm{WP}} W
    - \left\langle \Delta^{\mathrm{WP}} W \right\rangle_{\Xi} \right\|^2 \right\rangle_{\Xi}\\
    &\approx
    \left\langle \left\| \Delta^{\mathrm{WP}} W
    - \Delta^{\mathrm{SGD}} W \right\|^2 \right\rangle_{\Xi}.
    \end{aligned}
\end{equation}

Using the independence and fourth-moment assumptions on $\Xi$, this variance is approximated as follows (see, e.g., \citep{ren2023scaling, zuge_weight_2023}, for related derivations):
\begin{alignat}{1}
\mathrm{Var}_{\Xi}\!\left[\Delta^{\mathrm{WP}} W \mid z\right]
\approx
\eta^2
\left(\dim(\Xi)+\mu_{4,\Xi}-2\right)
\left\| \frac{\partial\ell(W;z)}{\partial W}\right\|^2,
\end{alignat}
where $\dim(\Xi)$ is the number of perturbed degrees of freedom, which is equal to $n_\mathrm{r}^2$ in the self-supervised setting of Section \ref{sec:sSL-ESN}.
Thus, WP can approximate SGD when the learning rate $\eta>0$ is sufficiently small for the perturbation-induced fluctuations inherited from $\Xi$ to be averaged out over successive updates.

Node perturbation (NP) \citep{flower_summed_1992, fiete_gradient_2006} follows a similar principle to WP, but uses lower-dimensional perturbations by exploiting the structure of the derivative to be approximated.
NP assumes that the instantaneous loss $\ell(W;z)$ has the following form:
\begin{equation}
        \ell(W;z)=\ell(y),
    \qquad
    y=Wz.
\end{equation}
This assumption naturally arises when $W$ and $y$ represent network connections and their outputs, respectively.
Under this assumption, the derivative in SGD has the following structure:
\begin{alignat}{1}
    \Delta^{\mathrm{SGD}} W
    =
    -\eta \frac{\partial \ell(W;z)}{\partial W}
    =
    -\eta \frac{\partial \ell(y)}{\partial y}z^\top.
\end{alignat}
NP approximates the derivative $\frac{\partial \ell(y)}{\partial y}$ in SGD as follows:
\begin{alignat}{1}
    \label{eq:NP}
    \Delta^{\mathrm{NP}} W
    =
    -\eta
    \frac{\ell(y+\alpha\xi)-\ell(y-\alpha\xi)}{2\alpha}
    \xi z^\top,
\end{alignat}
where $\xi$ is a perturbation vector of the same size as $y=Wz$.
We assume that its entries are independent, zero-mean, unit-variance random variables with a common bounded fourth moment:
\begin{equation}
     \left\langle \xi_i \right\rangle_\xi = 0,
     \qquad
     \left\langle \xi_i\xi_j \right\rangle_\xi = \delta_{ij},
     \qquad
     \left\langle \xi_i^4 \right\rangle_\xi = \mu_{4,\xi}<\infty .
\end{equation}
For sufficiently small $\alpha$, NP approximates the SGD update in expectation over the perturbation by the same argument as Eq.~\eqref{eq:WP-SGD}:
\begin{alignat}{1}
\label{eq:NP-SGD}
\left\langle \Delta^{\mathrm{NP}} W \right\rangle_{\xi}
\approx
\Delta^{\mathrm{SGD}} W.
\end{alignat}
The cost of this derivative-free approximation is the random deviation from SGD induced by $\xi$ for each fixed $z$, which is quantified by the corresponding estimation variance:
\begin{equation}
    \begin{aligned}
    \mathrm{Var}_{\xi}\!\left[\Delta^{\mathrm{NP}} W \mid z\right]
    &:=
    \left\langle \left\| \Delta^{\mathrm{NP}} W
    - \left\langle \Delta^{\mathrm{NP}} W \right\rangle_{\xi} \right\|^2 \right\rangle_{\xi}\\
    &\approx
    \left\langle \left\| \Delta^{\mathrm{NP}} W
    - \Delta^{\mathrm{SGD}} W \right\|^2 \right\rangle_{\xi}.
    \end{aligned}
\end{equation}
Using the independence and fourth-moment assumptions on $\xi$, this variance is approximated as follows (see, e.g., \citep{ren2023scaling, zuge_weight_2023}, for related derivations):
\begin{alignat}{1}
\mathrm{Var}_{\xi}\!\left[\Delta^{\mathrm{NP}} W \mid z\right]
\approx
\eta^2
\left(\dim(\xi)+\mu_{4,\xi}-2\right)
\left\| \frac{\partial \ell(y)}{\partial y}z^\top \right\|^2,
\end{alignat}
where $\dim(\xi)$ is the number of perturbed degrees of freedom, which is equal to $n_\mathrm{r}$ in the self-supervised setting of Section \ref{sec:sSL-ESN}.
Thus, NP can approximate SGD when the learning rate $\eta>0$ is sufficiently small for the perturbation-induced fluctuations inherited from $\xi$ to be averaged out over successive updates.

Consequently, the deviation of WP or NP from GD arises from two sources: the intrinsic SGD variance inherited from the randomness of $z$, and the additional estimation variance inherited from the perturbation.
For WP, this additional variance scales as $O(\eta^2 n_\mathrm{r}^2)$, whereas for NP it scales as $O(\eta^2 n_\mathrm{r})$.
Therefore, as $n_\mathrm{r}$ increases, WP requires a much smaller learning rate or more averaging to suppress the perturbation-induced fluctuations, while NP provides a lower-variance approximation to SGD by exploiting the structural knowledge that the loss evaluates the output $y=Wz$, rather than the parameter matrix $W$ itself.

\subsection{Self-Supervised Formulation of Echo State Networks}
\label{sec:sSL-ESN}
An echo state network (ESN) \citep{jaeger2001echo} consists of an input \(s_t\in\mathbb{R}^{n_{\mathrm{in}}}\), a reservoir state \(r_t\in\mathbb{R}^{n_\mathrm{r}}\), and an output \(o_t\in\mathbb{R}^{n_{\mathrm{out}}}\), where \(t\) denotes discrete time.
The ESN dynamics are given by
\begin{equation}
    \label{eq:ESN}
    \begin{aligned}
      r_{t+1} & = \sigma(\Win s_t + \Wr r_t), \\
      o_t     & = \Wout r_t,
    \end{aligned}
\end{equation}
where \(\sigma:\mathbb{R}^{n_\mathrm{r}}\to\mathbb{R}^{n_\mathrm{r}}\) is a nonlinear activation function, and \(\Win \in\mathbb{R}^{n_\mathrm{r}\times n_{\mathrm{in}}}\), \(\Wr \in\mathbb{R}^{n_\mathrm{r}\times n_\mathrm{r}}\), and \(\Wout \in\mathbb{R}^{n_{\mathrm{out}}\times n_\mathrm{r}}\) are the input, reservoir, and output connection matrices, respectively.
\(\Win \) and \(\Wr \) are randomly generated and fixed.
In the standard ESN setting \citep{jaeger2001echo}, only \(\Wout \) is trained so that the ESN output \(o_t\) approximates a desired output \(o_t^*\in\mathbb{R}^{n_{\mathrm{out}}}\), which is usually provided externally.
The optimal readout minimizing the squared loss is given by
\begin{alignat}{1}
  \label{eq:Wout*}
  \Wout _* = \langle o_t^* r_t^\top\rangle_{o_t,r_t}\langle r_t r_t^\top\rangle_{r_t}^{-1},
\end{alignat}
where we assume that the covariance matrix $\langle r_tr_t^\top\rangle_{r_t}$ is invertible.
Since this solution explicitly depends on the desired output \(o_t^*\), this formulation is supervised learning.
An additional mechanism is needed if the ESN is to learn autonomously without explicit external target signals.

In the special case where the desired output is the input,
\begin{alignat}{1}
  o_t^* = s_t,
\end{alignat}
a self-supervised formulation is possible \citep{yamada_unsupervised_2026}.
We introduce an artificial output \(\hat{o}_t\in\mathbb{R}^{n_\mathrm{r}}\) defined by
\begin{alignat}{1}
  \hat{o}_t = \Wdyn r_t,
\end{alignat}
where \(\Wdyn\in\mathbb{R}^{n_\mathrm{r}\times n_\mathrm{r}}\) is an artificial readout matrix.
We assume that the activation function \(\sigma\) is invertible, so that \(\sigma^{-1}\) exists.
The artificial output \(\hat{o}_t\) is trained to approximate \(\sigma^{-1}(r_{t+1})\), which is determined internally by the ESN dynamics.
The optimal solution minimizing the squared loss is
\begin{alignat}{1}
  \label{eq:Wdyn*}
  \Wdyn_* = \langle \sigma^{-1}(r_{t+1}) r_t^\top\rangle_{r_t,r_{t+1}}\langle r_t r_t^\top\rangle_{r_t}^{-1},
\end{alignat}
where again we assume that the covariance matrix $\langle r_tr_t^\top\rangle_{r_t}$ is invertible.
This solution requires only reservoir states and does not explicitly use the input \(s_t\) as the desired output.
Thus, the training of \(\Wdyn\) can be regarded as self-supervised learning, with \(\sigma^{-1}(r_{t+1})\) providing an internally available target.

The self-supervised learning of \(\Wdyn\) is related to supervised learning with \(o_t^*=s_t\) as follows.
Define the transformation \(\mathcal{P}\) by
\begin{alignat}{1}
  \label{eq:P}
  \mathcal{P}(\Wout ) := \Win \Wout +\Wr .
\end{alignat}
Since \(\Win \) and \(\Wr \) are randomly generated and fixed, the transformation \(\mathcal{P}\) is static; that is, it can be applied without knowing the dynamic variables \(s_t\), \(r_t\), and \(o_t^*\).
Using the definitions of the optimal solutions \eqref{eq:Wout*} and \eqref{eq:Wdyn*}, together with the ESN dynamics \eqref{eq:ESN}, we obtain
\begin{alignat}{1}
  \label{eq:Wout-Wdyn}
  \Wdyn_* = \mathcal{P}(\Wout _*).
\end{alignat}
Hence, the optimal self-supervised readout \(\Wdyn_*\) is obtained from the optimal supervised readout \(\Wout _*\) by applying the static transformation \(\mathcal{P}\).

Conversely, suppose that $\Win$ has full column rank. 
This requires $n_{\mathrm{in}} \leq n_{\mathrm{r}}$, which we assume throughout this paper.
Then \((\Win )^+\Win =I_{n_{\mathrm{in}}}\), where \((\Win )^+\) denotes the Moore--Penrose inverse of \(\Win \).
Thus, the left inverse \(\mathcal{Q}\) of the transformation \(\mathcal{P}\) exists and is defined by
\begin{alignat}{1}
  \label{eq:Q}
  \mathcal{Q}(\Wdyn) := (\Win )^+(\Wdyn-\Wr ).
\end{alignat}
The transformation \(\mathcal{Q}\) is static, as is \(\mathcal{P}\), and satisfies \(\mathcal{Q}\circ\mathcal{P}=\mathrm{id}\).
Thus, applying the transformation \(\mathcal{Q}\) to both sides of \eqref{eq:Wout-Wdyn} yields
\begin{alignat}{1}
  \mathcal{Q}(\Wdyn_*) = \Wout _*.
\end{alignat}
Therefore, supervised learning of \(\Wout \) in ESNs with \(s_t\) as the desired output \(o_t^*\) can be reformulated as self-supervised learning of \(\Wdyn\), followed by the static transformation \(\mathcal{Q}\).
In this paper, we focus on the self-supervised learning of \(\Wdyn\) instead of the supervised learning of \(\Wout \).

Online learning rules for $\Wdyn$ are derived from minimization algorithms for the cost function
\begin{equation}
    L^{\mathrm{sSL}}(\Wdyn)
    =
    \left\langle
    \ell^{\mathrm{sSL}}(\Wdyn;r_t,r_{t+1})
    \right\rangle_{r_t,r_{t+1}},
\end{equation}
where
\begin{alignat}{1}
  \label{eq:sSL-cost}
  \ell^{\mathrm{sSL}}(\Wdyn;r_t,r_{t+1})
  =
  \frac{1}{2}\|\Wdyn r_t-\sigma^{-1}(r_{t+1})\|^2.
\end{alignat}
Note that this cost function does not explicitly depend on the supervised signal $o_t^*=s_t$, and neither do the corresponding online learning rules.

As reviewed in Section~\ref{sec:learning-alg}, applying SGD, WP, and NP to a given instantaneous loss yields sample-wise online update rules.
To specify that the instantaneous loss is $\ell^{\mathrm{sSL}}$, we refer to these rules as
$\mathrm{SGD}(\ell^{\mathrm{sSL}})$,
$\mathrm{WP}(\ell^{\mathrm{sSL}})$, and
$\mathrm{NP}(\ell^{\mathrm{sSL}})$, respectively.
With the shorthand
$y_t=\Wdyn r_t$, and by a slight abuse of notation
$\ell^{\mathrm{sSL}}(\Wdyn;r_t,r_{t+1})
=
\ell^{\mathrm{sSL}}(y_t;r_{t+1})$,
the corresponding updates are given by
\begin{alignat}{1}
  \label{eq:SGD-Wdyn}
  \Delta^{\mathrm{SGD}} \Wdyn
  &=
  -\eta
  \frac{\partial \ell^{\mathrm{sSL}}(\Wdyn;r_t,r_{t+1})}{\partial \Wdyn}
  =
  -\eta
  \frac{\partial \ell^{\mathrm{sSL}}(y_t;r_{t+1})}{\partial y_t}
  r_t^\top,\\
  \label{eq:WP-Wdyn}
  \Delta^{\mathrm{WP}} \Wdyn
  &=
  -\eta
  \frac{
  \ell^{\mathrm{sSL}}(\Wdyn+\alpha\Xi;r_t,r_{t+1})
  -
  \ell^{\mathrm{sSL}}(\Wdyn-\alpha\Xi;r_t,r_{t+1})
  }{2\alpha}\Xi,\\
  \label{eq:NP-Wdyn}
  \Delta^{\mathrm{NP}} \Wdyn
  &=
  -\eta
  \frac{
  \ell^{\mathrm{sSL}}(y_t+\alpha\xi;r_{t+1})
  -
  \ell^{\mathrm{sSL}}(y_t-\alpha\xi;r_{t+1})
  }{2\alpha}
  \xi r_t^\top.
\end{alignat}
SGD$(\ell^{\mathrm{sSL}})$, WP$(\ell^{\mathrm{sSL}})$, and NP$(\ell^{\mathrm{sSL}})$ are valid in principle, as reviewed in Section~\ref{sec:learning-alg}, but they face practical limitations as the reservoir dimension $n_{\mathrm{r}}=\dim r_t$ increases.
Specifically, SGD$(\ell^{\mathrm{sSL}})$ requires handling the error vector \(\partial \ell^{\mathrm{sSL}}(y_t;r_{t+1})/\partial y_t\), whose size is \(n_\mathrm{r}\), and therefore requires \(O(n_\mathrm{r})\) memory and routing costs.
Although the perturbation-based methods WP$(\ell^{\mathrm{sSL}})$ and NP$(\ell^{\mathrm{sSL}})$ alleviate these costs by representing the error feedback as a scalar, they instead suffer from a variance problem because they introduce artificial perturbations \(\Xi\) and \(\xi\) of sizes \(n_\mathrm{r}^2\) and \(n_\mathrm{r}\), respectively.
Consequently, these online learning algorithms are not suitable for implementation on edge devices with large reservoirs.
Therefore, the remaining question is whether one can retain scalar error feedback while reducing the effective perturbation dimension from \(n_\mathrm{r}\) to a smaller dimension.
In the next section, we show that the orthogonal structure of \(\ell^{\mathrm{sSL}}\) allows such a reduction.

\subsection{Orthogonal Cost Decomposition and Learning Algorithm}
\label{sec:proposed}
We show that the self-supervised cost \(\ell^{\mathrm{sSL}}(\Wdyn;r_t,r_{t+1})\) admits the following orthogonal decomposition:
\begin{alignat}{1}
\label{eq:OD}
  \ell^{\mathrm{sSL}}(\Wdyn;r_t,r_{t+1})
  &=
  \tilde{\ell}^{\mathrm{sSL}}
  (\Wdyn;r_t,r_{t+1})
  +
  \frac{1}{2}
  \|\Pi_\perp(\Wdyn)\|^2_{I,r_t r_t^\top},
\end{alignat}
where
\begin{alignat}{1}
  \tilde{\ell}^{\mathrm{sSL}}
  (\Wdyn;r_t,r_{t+1})
  &=
  \frac{1}{2}
  \left\|
  (\Win )^+
  \left[
    \Wdyn r_t-\sigma^{-1}(r_{t+1})
  \right]
  \right\|^2_{(\Win )^\top \Win }, \\
  \Pi_\perp(\Wdyn)
  &=
  \left(I-\Win (\Win )^+\right)
  (\Wdyn-\Wr).
\end{alignat}
The decomposition in Eq.~(\ref{eq:OD}) follows directly from the Pythagorean decomposition of the norm with respect to the orthogonal direct sum $\mathbb{R}^{n_{\mathrm{r}}} = \operatorname{Im}(\Win)\oplus \operatorname{Im}(\Win)^\perp$.
\begin{revision}
See \ref{app:OD} for the proof.
\end{revision}

The decomposition (\ref{eq:OD}) leads to two observations.
First, the first term of Eq.~(\ref{eq:OD}) depends on \(\Wdyn\) only through
\begin{alignat}{1}
  \tilde{y}_t
  =
  (\Win )^+\Wdyn r_t
  \in \mathbb{R}^{n_{\mathrm{in}}}.
\end{alignat}
Thus, we may write
\begin{alignat}{1}
  \tilde{\ell}^{\mathrm{sSL}}
  (\Wdyn;r_t,r_{t+1})
  =
  \tilde{\ell}^{\mathrm{sSL}}
  (\tilde{y}_t;r_{t+1}).
\end{alignat}
This implies the following structure of the derivative of $\tilde{\ell}^{\mathrm{sSL}}$ with respect to $\Wdyn$:
\begin{alignat}{1}
    \label{eq:del-tilde-ell}
    \frac{\partial \tilde{\ell}^{\mathrm{sSL}}}{\partial \Wdyn} = ((\Win)^+)^\top \frac{\partial \tilde{\ell}^{\mathrm{sSL}}}{\partial \tilde{y}_t}r_t^\top.
\end{alignat}
By applying an argument analogous to node perturbation, we obtain the following perturbation-based online learning rule:
\begin{alignat}{1}
  \label{eq:Ours}
  \tilde{\Delta}^{\mathrm{Ours}} \Wdyn
  =
  -\eta
  \frac{
  \tilde{\ell}^{\mathrm{sSL}}
  (\tilde{y}_t+\alpha\tilde{\xi};r_{t+1})
  -
  \tilde{\ell}^{\mathrm{sSL}}
  (\tilde{y}_t-\alpha\tilde{\xi};r_{t+1})
  }{2\alpha}
  \bigl((\Win )^+\bigr)^\top
  \tilde{\xi} r_t^\top,
\end{alignat}
where $\tilde{\xi}\in\mathbb{R}^{n_{\mathrm{in}}}$ is a perturbation vector whose entries are independent, zero-mean, unit-variance random variables with a common bounded fourth moment:
\begin{align}
    \left\langle \tilde{\xi}_i \right\rangle_{\tilde{\xi}} &= 0,
    &
    \left\langle \tilde{\xi}_i\tilde{\xi}_j \right\rangle_{\tilde{\xi}}
    &= \delta_{ij},
    &
    \left\langle \tilde{\xi}_i^4 \right\rangle_{\tilde{\xi}}
    &= \mu_{4,\tilde{\xi}} < \infty.
    \label{eq:proposed-perturbation-assumption}
\end{align}
As with WP and NP, the proposed rule approximates SGD with respect to
\(\tilde{\ell}^{\mathrm{sSL}}\) for sufficiently small \(\alpha\):
\begin{equation}
    \label{eq:Ours-SGD}
    \begin{aligned}
      \left\langle
      \tilde{\Delta}^{\mathrm{Ours}} \Wdyn
      \right\rangle_{\tilde{\xi}}
      &\approx
      -\eta
      ((\Win )^+)^\top \frac{\partial \tilde{\ell}^{\mathrm{sSL}}}{\partial \tilde{y}_t}r_t^\top \\
      &=
      -\eta
      \frac{\partial
      \tilde{\ell}^{\mathrm{sSL}}
      (\Wdyn;r_t,r_{t+1})}
      {\partial \Wdyn}.
    \end{aligned}
\end{equation}

Unlike WP and NP, whose perturbation dimensions are $n_\mathrm{r}^2$ and $n_\mathrm{r}$ respectively, the proposed rule (\ref{eq:Ours}) uses a perturbation vector of dimension \(n_{\mathrm{in}}\).
Thus, the dimension-dependent factor in the perturbation-induced variance scales as \(O(n_{\mathrm{in}})\), rather than \(O(n_\mathrm{r})\).

Second, the second term of Eq.~(\ref{eq:OD}) can be removed without using the data \(r_t\), and without changing the first term.
Define the static transformation
\begin{alignat}{1}
  \mathcal{S}(\Wdyn)
  =
  \Win (\Win )^+\Wdyn
  +
  \left(I-\Win (\Win )^+\right)\Wr  .
\end{alignat}
The static transformation $\mathcal{S}$ satisfies
\begin{alignat}{1}
  \tilde{\ell}^{\mathrm{sSL}}(\mathcal{S}(\Wdyn);r_t,r_{t+1})&=\tilde{\ell}^{\mathrm{sSL}}(\Wdyn;r_t,r_{t+1}), \\
  \Pi_\perp\circ\mathcal{S}(\Wdyn) &= O.
\end{alignat}
Therefore, during online learning from sequentially given reservoir states, the second term of Eq.~(\ref{eq:OD}) can be eliminated offline without affecting the online training progress associated with the first term.
In the proposed method, we therefore update \(\Wdyn\) using the online rule for \(\tilde{\ell}^{\mathrm{sSL}}\), and apply \(\mathcal{S}\) to the current \(\Wdyn\) at any time during learning.
Consequently, the proposed method provides an online self-supervised learning rule for ESNs whose error-transmission pathway size and perturbation dimension do not scale with the reservoir dimension.
This property makes the method suitable for high-dimensional reservoirs and hardware implementations in edge environments.

The key implication of the orthogonal decomposition above is that, for online adaptation of $\Wdyn$, it is sufficient to minimize the reduced loss $\tilde{\ell}^{\mathrm{sSL}}$ rather than the original self-supervised loss $\ell^{\mathrm{sSL}}$. The dependence of $\tilde{\ell}^{\mathrm{sSL}}$ on $\Wdyn$ is mediated by the nested map
\begin{equation}
    \Wdyn
    \longmapsto
    y_t=\Wdyn r_t
    \longmapsto
    \tilde{y}_t=(\Win)^+y_t.
\end{equation}
This nested structure induces the following hierarchy of applicability for perturbation-based learning rules:
\begin{equation}
    \mathrm{SGD}(\tilde{\ell}^{\mathrm{sSL}}),\ \mathrm{WP}(\tilde{\ell}^{\mathrm{sSL}})
    \;\longleftarrow\;
    \mathrm{NP}(\tilde{\ell}^{\mathrm{sSL}})
    \;\longleftarrow\;
    \mathrm{Ours}.
\end{equation}
Here, the arrows indicate logical implication in applicability: if a rule on the right is well defined, then every rule to its left is also well defined. $\mathrm{SGD}(\tilde{\ell}^{\mathrm{sSL}})$ and $\mathrm{WP}(\tilde{\ell}^{\mathrm{sSL}})$ operate directly in parameter space and require only access to the reduced loss as a function of $\Wdyn$. $\mathrm{NP}(\tilde{\ell}^{\mathrm{sSL}})$ additionally exploits the fact that the reduced loss can be evaluated through the reservoir-dimensional output $y_t=\Wdyn r_t$. The proposed rule further exploits the redundancy that $\tilde{\ell}^{\mathrm{sSL}}$ depends on $y_t$ only through the lower-dimensional variable $\tilde{y}_t$.

Consequently, whenever the proposed rule is well defined, $\mathrm{NP}(\tilde{\ell}^{\mathrm{sSL}})$ is also well defined; whenever $\mathrm{NP}(\tilde{\ell}^{\mathrm{sSL}})$ is well defined, $\mathrm{SGD}(\tilde{\ell}^{\mathrm{sSL}})$ and $\mathrm{WP}(\tilde{\ell}^{\mathrm{sSL}})$ are also well defined. These reduced-loss rules should be distinguished from the vanilla rules $\mathrm{SGD}(\ell^{\mathrm{sSL}})$, $\mathrm{WP}(\ell^{\mathrm{sSL}})$, and $\mathrm{NP}(\ell^{\mathrm{sSL}})$ defined in Eqs.~\eqref{eq:SGD-Wdyn}, \eqref{eq:WP-Wdyn}, and \eqref{eq:NP-Wdyn}, respectively, because the reduced-loss rules explicitly exploit the orthogonal decomposition of the original loss.

The detailed relationships among these rules are analyzed in \ref{app:perturbation-relations}. In particular, the perturbation rules constructed from the reduced loss reproduce the corresponding reduced-loss SGD update in expectation, whereas their differences at individual perturbation realizations are characterized by additional zero-mean components. The proposed rule also admits a canonical-dual NP interpretation driven by the lower-dimensional perturbation $\tilde{\xi}\in\mathbb{R}^{n_{\mathrm{in}}}$.

Because $\mathrm{SGD}(\tilde{\ell}^{\mathrm{sSL}})$, $\mathrm{WP}(\tilde{\ell}^{\mathrm{sSL}})$, and $\mathrm{NP}(\tilde{\ell}^{\mathrm{sSL}})$ already exploit the orthogonal decomposition introduced in this work, we use them as analytical reference rules rather than treating them as independent empirical baselines. 
Accordingly, the numerical experiments compare the proposed method only with the vanilla rules $\mathrm{SGD}(\ell^{\mathrm{sSL}})$, $\mathrm{WP}(\ell^{\mathrm{sSL}})$, and $\mathrm{NP}(\ell^{\mathrm{sSL}})$.
\section{Numerical Examples}
\label{sec:numerical}

We numerically demonstrate the features of the proposed perturbation-based online learning algorithm for self-supervised learning in ESNs.

\begin{figure}[H]
\begin{center}
\includegraphics[width=\linewidth]{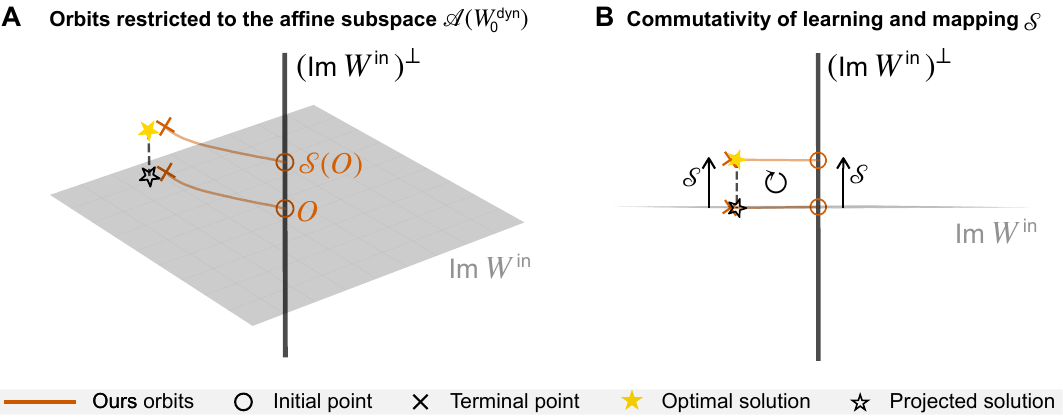}
\caption{Numerically computed learning orbits of the proposed rule, illustrating the geometric interpretation from the orthogonal decomposition of the self-supervised loss (\(n_\mathrm{r}=3\)). Orbits are shown relative to the input subspace \(\mathrm{Im}\,W^{\mathrm{in}}\) and its orthogonal complement, each running from its initial point (circle) to its terminal point (cross). (A) Since each column is updated only within the input subspace, every orbit is confined to an affine subspace through its initial value. The orbit from \(O\) reaches the projected solution and the orbit from \(\mathcal{S}(O)\) reaches the optimal solution; the two coincide within the input subspace and differ only along its orthogonal complement (dashed line). (B) The transformation \(\mathcal{S}\) acts only along the orthogonal complement while online learning acts only within the input subspace, so the two commute (\(\circlearrowright\)): applying \(\mathcal{S}\) before or after learning reaches the same optimal solution.}
\label{fig:orbit}
\end{center}
\end{figure}

The orthogonal decomposition of \(\ell^{\mathrm{sSL}}\) yields a geometric interpretation of the learning dynamics (Figure~\ref{fig:orbit}). We plot the online orbits of the proposed method obtained from the numerical simulations; see \ref{app:simulation} for the detailed setting. For visualization we set \(n_\mathrm{r}=3\) and show the trajectory of the first column of the \(3\times 3\) matrix \(W^{\mathrm{dyn}}\), from its initial point (\(\circ\)) to its terminal point (\(\times\)).

Whereas the full cost \(\ell^{\mathrm{sSL}}(\Wdyn)\) is minimized at the optimal solution (the target of full-cost methods such as SGD, WP, and NP), the proposed method optimizes only its projected component \(\tilde{\ell}^{\mathrm{sSL}}(\Wdyn)\) using the gradient in Eq.~(\ref{eq:del-tilde-ell}). Since the update direction of each column belongs to \(\operatorname{Im}W^{\mathrm{in}}\), the orbit is confined to the affine subspace
\[
  \mathcal{A}(W^{\mathrm{dyn}}_0)
  := W^{\mathrm{dyn}}_0 + \mathcal{M}_{\mathrm{in}},
  \qquad
  \mathcal{M}_{\mathrm{in}}
  := \{\, W^{\mathrm{in}}Z \mid Z\in\mathbb{R}^{n_{\mathrm{in}}\times n_\mathrm{r}} \,\},
\]
where \(\mathcal{M}_{\mathrm{in}}\) is the matrix-valued extension of \(\operatorname{Im}W^{\mathrm{in}}\); for \(W^{\mathrm{dyn}}_0=O\) it reduces to the linear subspace \(\mathcal{M}_{\mathrm{in}}\) (Figure~\ref{fig:orbit}A). Starting from \(W^{\mathrm{dyn}}_0=O\), the orbit therefore stays in \(\operatorname{Im}W^{\mathrm{in}}\) and converges to the projected solution, i.e., the projection of the optimal solution onto \(\operatorname{Im}W^{\mathrm{in}}\). The two solutions coincide within \(\operatorname{Im}W^{\mathrm{in}}\) and differ only along the orthogonal complement \(\mathcal{M}_{\mathrm{in}}^{\perp}\) (dashed line).

This residual component along \(\mathcal{M}_{\mathrm{in}}^{\perp}\) is never updated online. Instead, the proposed method resolves the corresponding orthogonal (offline) term \(\|\Pi_\perp(\Wdyn)\|^2_{I,r_t r_t^\top}/2\) by the static transformation \(\mathcal{S}\), which requires no reservoir states \(r_t\). Because \(\mathcal{S}\) acts only on \(\mathcal{M}_{\mathrm{in}}^{\perp}\) while online learning acts only within \(\mathcal{M}_{\mathrm{in}}\), the two operations commute (Figure~\ref{fig:orbit}B). Hence \(\mathcal{S}\) may be applied at any point during learning: applying it after learning (starting from \(W^{\mathrm{dyn}}_0=O\)) and applying it before learning (starting from the residual-free initial value \(W^{\mathrm{dyn}}_0=\mathcal{S}(O)\)) both reach the same optimal solution.

\begin{figure}[H]
\begin{center}
\includegraphics[width=\linewidth]{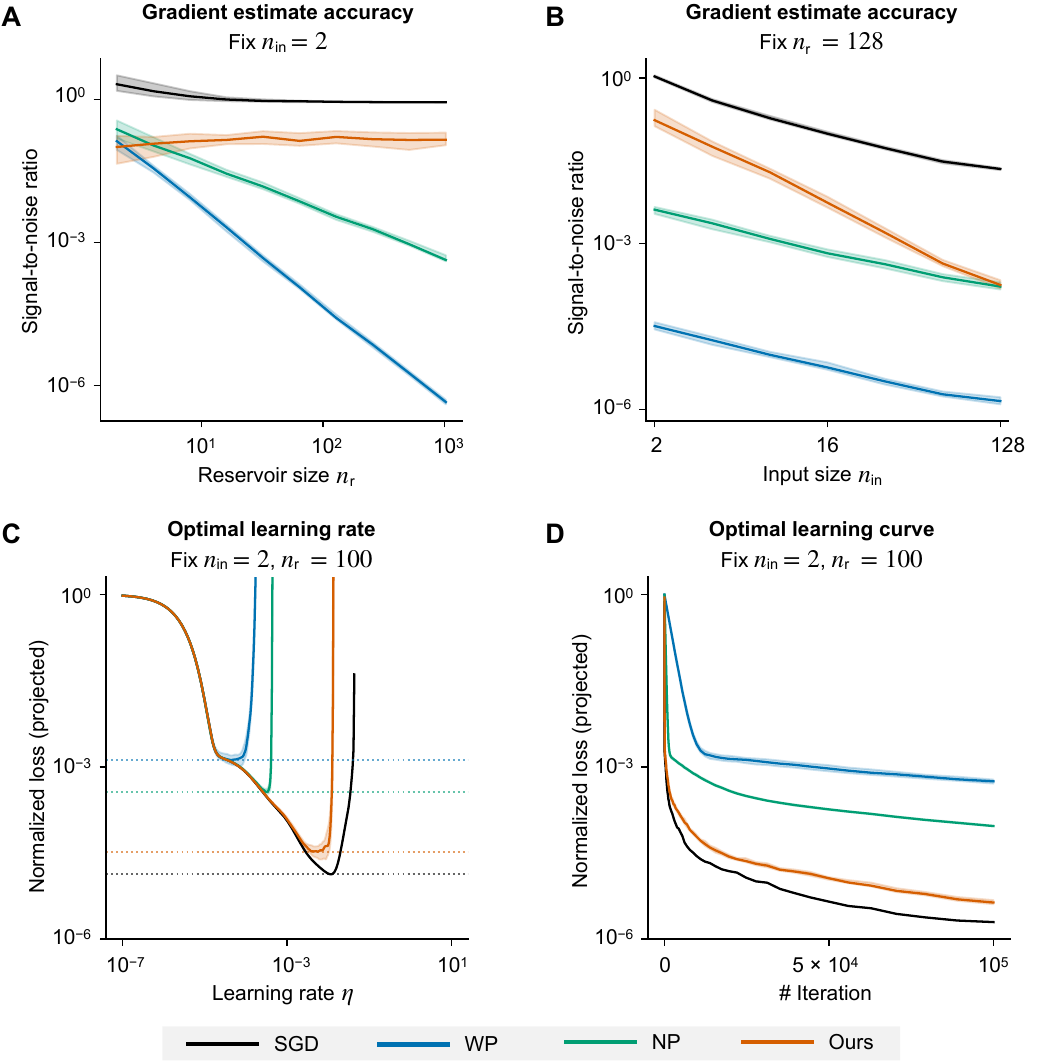}
\caption{
Performance evaluation of the proposed perturbation-based online learning algorithm.
(A) Signal-to-noise ratio (SNR) of the update direction versus the reservoir size \(n_\mathrm{r}\) (\(n_{\mathrm{in}} = 2\)).
The SNRs of WP and NP deteriorate with \(n_\mathrm{r}\), whereas Ours remains constant because its perturbation dimension is \(n_{\mathrm{in}}\), not \(n_\mathrm{r}\).
(B) SNR versus the input dimension $n_{\mathrm{in}}$ at fixed $n_\mathrm{r}=128$.
For SGD, WP, and NP, the decrease mainly reflects the change in the signal
term with the task, whereas for Ours it additionally reflects the
$n_{\mathrm{in}}$-dependent perturbation-induced variance. Consequently, the
SNR advantage of Ours over NP decreases as $n_{\mathrm{in}}$ approaches $n_\mathrm{r}$
and vanishes at $n_{\mathrm{in}}=n_\mathrm{r}$ under the orthonormal-$W^{\mathrm{in}}$
and Gaussian-perturbation setting.
(C) Learning-rate dependence of the final normalized loss (\(n_{\mathrm{in}} = 2\), \(n_\mathrm{r} = 100\)); dotted horizontal lines mark the minimum for each algorithm.
(D) Learning curves at the optimal learning rates from panel C.
Lines and shaded bands show the median and interquartile range over random realizations.
}
\label{fig:eval}
\end{center}
\end{figure}

Figure~\ref{fig:eval} evaluates the proposed method (Ours) in comparison with SGD, WP, and NP in terms of reservoir-size dependence, input-size dependence, learning-rate sensitivity, and online learning dynamics.
As reviewed in Section~\ref{sec:learning-alg}, perturbation-based learning algorithms, including WP, NP, and Ours, approximate SGD on average by estimating the gradient from scalar-valued error signals using stochastic perturbations.
These stochastic perturbations introduce additional variance into the update rules.
Therefore, perturbation-based learning algorithms introduce additional variance into their update estimates relative to their respective reference gradients. In the present experiments, this additional variance results in lower signal-to-noise ratios (SNRs; defined in Eq.~(\ref{eq:SNR})) than for SGD (Figure~\ref{fig:eval}A).
Here, we focus on how this SNR degradation depends on the reservoir size.
Since the additional variance scales with the perturbation dimension, the SNRs of WP and NP deteriorate as the reservoir size \(n_\mathrm{r}\) increases (Figure~\ref{fig:eval}A, blue and green lines).
In contrast, the perturbation dimension of Ours scales with the input dimension \(n_{\mathrm{in}}\), rather than with \(n_\mathrm{r}\) (Figure~\ref{fig:eval}A, orange line).
Consequently, the SNR of Ours remains approximately independent of \(n_\mathrm{r}\), indicating that the proposed method is scalable to large reservoirs.

Figure~\ref{fig:eval}B examines the complementary dependence on the input dimension
$n_{\mathrm{in}}$ at a fixed reservoir size. Although the raw SNRs of all four
rules decrease as $n_{\mathrm{in}}$ increases, Figure~S1 shows that the underlying
causes differ. For SGD, WP, and NP, the decrease mainly reflects a common
reduction in the signal term caused by the change in the task with
$n_{\mathrm{in}}$; because $n_\mathrm{r}$ is fixed, the perturbation-induced variances of
WP and NP remain approximately independent of $n_{\mathrm{in}}$. For Ours, by
contrast, the perturbation-induced variance also increases with
$n_{\mathrm{in}}$, because its perturbation source has dimension
$n_{\mathrm{in}}$. Consequently, the SNR advantage of Ours over NP is largest
when $n_{\mathrm{in}}\ll n_\mathrm{r}$ and decreases as $n_{\mathrm{in}}$ approaches
$n_\mathrm{r}$. Under the orthonormal-$W^{\mathrm{in}}$ and Gaussian-perturbation setting
used here, the gap vanishes at $n_{\mathrm{in}}=n_\mathrm{r}$.

Reducing the learning rate generally improves stability and suppresses stochastic fluctuations.
In practice, however, the number of iterations is finite, so an excessively small learning rate may result in insufficient progress within the available training horizon.
On the other hand, because the gradient gives only a local descent direction and its online estimate is noisy, an excessively large learning rate can cause divergence of the parameters.
Therefore, for a fixed finite number of iterations, the choice of learning rate involves a finite-horizon trade-off, and the learning rate is selected empirically here.
A larger SNR implies that each update is better aligned with the reference gradient, allowing efficient learning with a larger learning rate while avoiding divergence.
The numerical results in Figure~\ref{fig:eval}C show that the proposed method attains a lower normalized loss over a broader range of learning rates than WP and NP.
This suggests that reducing the effective perturbation dimension improves the practical stability and efficiency of perturbation-based online learning.
The learning curves in Figure~\ref{fig:eval}D, obtained using the learning rates selected from Figure~\ref{fig:eval}C, further confirm the effectiveness of the proposed method for large reservoirs.
Ours converges faster and more stably than WP and NP, while retaining the scalar-valued error feedback that characterizes perturbation-based learning.

\section{Discussion}
\label{sec:discussion}

In this study, we considered learning algorithms for ESNs that satisfy three requirements: autonomous adaptation through self-supervised learning, sequential adaptation through online learning, and scalability to large reservoirs.
Building on the self-supervised formulation of ESNs proposed in our previous work \citep{yamada_unsupervised_2026}, we compared three existing online learning rules: SGD, WP, and NP.
We argued that SGD requires the transmission and storage of a high-dimensional error vector, whose memory and routing costs scale with the reservoir size.
Perturbation-based methods such as WP and NP avoid this difficulty by representing the error feedback as a scalar loss difference.
Such perturbation-based learning has attracted attention as one possible approach to local learning that estimates gradients without explicit error backpropagation \citep{jabri_weight_1991, flower_summed_1992, fiete_gradient_2006}.
However, this scalarization comes at the cost of additional variance in the online gradient estimate, and this variance increases with the dimension of the perturbation.
Since the perturbation dimensions of WP and NP scale with the reservoir size, their gradient estimates become increasingly noisy for large reservoirs, limiting their scalability \citep{werfel_learning_2005}.
To overcome this trade-off, we proposed a new perturbation-based online learning rule that inherits the advantages of both approaches.
Because the proposed method is perturbation-based, it uses scalar-valued error feedback and therefore avoids reservoir-size-dependent error-feedback memory and routing costs.
At the same time, by exploiting the specific structure of self-supervised learning in ESNs, the perturbation dimension is reduced to the input dimension, rather than the reservoir dimension.
Consequently, the proposed method achieves scalar-valued error feedback without introducing reservoir-size-dependent perturbation variance.

Importantly, the proposed method does not eliminate the
reservoir-size dependence of the model itself:
$\Wdyn\in\mathbb{R}^{n_\mathrm{r}\times n_\mathrm{r}}$ and
$r_t\in\mathbb{R}^{n_\mathrm{r}}$ retain their original dimensions.
The scalability established here concerns the error-feedback pathway,
the perturbation-source dimension, and the resulting
perturbation-induced variance.
We also emphasize that the proposed rule relies on knowledge of the static structure $\Win$ and $\Wr$ through $\mathcal{S}$, not of the dynamic signal $s_t$: the fixed matrices are properties of the system itself and can in principle be characterized offline, whereas the external input is unavailable as a teaching signal by assumption \citep{yamada_unsupervised_2026}.
The remainder of this section examines the structural reason why such a reduction is possible.

\subsection{Redundancy of the self-supervised objective and its orthogonal decomposition}
\label{sec:disc-redundancy}

The key observation behind the proposed method is the orthogonal decomposition of the self-supervised cost function.
This decomposition separates the cost into an online component, which depends on reservoir states obtained sequentially through the input-driven dynamics, and an offline component, which can be computed from the fixed reservoir structure.
By optimizing only the online component during sequential learning and eliminating the offline component through a static transformation, the proposed method clarifies which part of the self-supervised objective should be allocated online computational resources.
This perspective provides a theoretical guideline for designing scalable online learning algorithms in resource-limited environments such as edge devices.

The mechanism behind this decomposition can be understood from the relation between supervised learning and self-supervised learning in ESNs.
In the input reconstruction task considered in this study, the supervised learning objective is redundantly embedded by the fixed input matrix \(W^{\mathrm{in}}\).
This embedding is justified by the full column rank assumption on \(W^{\mathrm{in}}\).
Indeed, if \(W^{\mathrm{in}}\) has full column rank, then it is injective, admits the left inverse \((W^{\mathrm{in}})^+\), and \((W^{\mathrm{in}})^\top W^{\mathrm{in}}\) is symmetric positive definite.
Therefore, \((W^{\mathrm{in}})^\top W^{\mathrm{in}}\) defines a valid metric on the input space, and the supervised error with metric \((W^{\mathrm{in}})^\top W^{\mathrm{in}}\) can be written equivalently as
\begin{equation}
    \| \underbrace{W^{\mathrm{out}}r_t}_{\mathrm{SL\ output}} - \underbrace{s_t}_{\mathrm{SL\ target}}\|^2_{(W^{\mathrm{in}})^\top W^{\mathrm{in}}} = \| \underbrace{W^{\mathrm{dyn}}r_t}_{\mathrm{sSL\ output}} - \underbrace{\sigma^{-1}(r_{t+1})}_{\mathrm{sSL\ target}}\|^2 .
\end{equation}

Thus, the self-supervised formulation can be regarded as a redundant representation of the supervised input reconstruction task induced by the injective map \(W^{\mathrm{in}}\).

Because this redundancy is introduced through \(W^{\mathrm{in}}\), the self-supervised error can be decomposed according to
\(\operatorname{Im}W^{\mathrm{in}}\oplus(\operatorname{Im}W^{\mathrm{in}})^\perp\).
Let \(P_{\mathrm{in}}\) and \(P_{\mathrm{in}}^\perp\) denote the orthogonal projections onto these two subspaces.
Then
\begin{equation}
\label{eq:disc-decomp}
    \begin{aligned}
      &\| W^{\mathrm{dyn}}r_t-\sigma^{-1}(r_{t+1})\|^2\\
      =&
      \| W^{\mathrm{dyn}}r_t-\sigma^{-1}(r_{t+1})\|^2_{P_{\mathrm{in}}}
      +
      \| W^{\mathrm{dyn}}r_t-\sigma^{-1}(r_{t+1})\|^2_{P_{\mathrm{in}}^\perp}\\
      =&
      \underbrace{\| W^{\mathrm{dyn}}r_t-\sigma^{-1}(r_{t+1})\|^2_{P_{\mathrm{in}}}}_{\text{low-dimensional online component}}
      +
      \underbrace{\| (W^{\mathrm{dyn}}-W^{\mathrm{rec}})r_t\|^2_{P_{\mathrm{in}}^\perp}}_{\text{offline residual component}} .
    \end{aligned}
\end{equation}
The first term is the component in \(\operatorname{Im}W^{\mathrm{in}}\), which contains the contribution of the external input \(s_t\).
The second term is the orthogonal component.
Since \(W^{\mathrm{in}}s_t\in\operatorname{Im}W^{\mathrm{in}}\), the direct contribution of \(s_t\) vanishes under \(P_{\mathrm{in}}^\perp\), and the residual reduces to \(\| (W^{\mathrm{dyn}}-W^{\mathrm{rec}})r_t\|^2_{P_{\mathrm{in}}^\perp}\), as shown in the second line of Eq.~\eqref{eq:disc-decomp}.
The online component corresponds to the redundancy-reduced part of the self-supervised objective and is directly related to the original supervised input reconstruction task.
In contrast, the residual component no longer contains the direct input term \(W^{\mathrm{in}}s_t\).
It is determined by the mismatch between \(W^{\mathrm{dyn}}\) and the fixed reservoir matrix \(W^{\mathrm{rec}}\) outside \(\operatorname{Im}W^{\mathrm{in}}\), and can be removed by the static transformation \(\mathcal{S}\).

This observation gives a broader interpretation of the proposed method.
When a supervised task is compatible with the system dynamics, it may be reformulated as a self-supervised task by introducing a redundant representation through fixed system matrices.
The validity of such a reformulation depends on whether the embedding preserves the original supervised error, as \(W^{\mathrm{in}}\) does here through its full column rank property.
The resulting self-supervised cost may contain components that are not essential for the original supervised task.
If the introduced redundancy is known, these redundant components can be identified by projection and removed from the online optimization problem.
Thus, self-supervised reformulation is not merely a way to construct internal targets; it can also reveal how online learning should be restricted to the dynamically necessary component of the objective.
This provides a principle for designing efficient perturbation-based learning algorithms whose error-feedback memory and routing costs, as well as perturbation-source dimension, depend on the input dimension rather than the reservoir dimension.

\subsection{Static transformations and the dynamical-system interpretation}
\label{sec:disc-static}

The affine map \(\mathcal{P}\) was originally introduced \citep{jaeger2001echo, lu_reservoir_2017, pathak_model-free_2018} to interpret the learned readout \(W^{\mathrm{out}}\) as a learned dynamical system. Specifically, the supervised learning result \(W^{\mathrm{out}}\) defines
\begin{alignat}{1}
    \hat{f}^{\mathrm{SL}}: r \mapsto \sigma\!\left[\mathcal{P}(W^{\mathrm{out}})r\right].
\end{alignat}
Previous studies have shown that the learned dynamical system \(\hat{f}^{\mathrm{SL}}\) can mimic the dynamics of the external input, \(s_t \mapsto s_{t+1}\) \citep{lu_reservoir_2017, pathak_model-free_2018, hara_learning_2022}. 
In addition, such learned dynamics can be combined with filtering algorithms based on state-space models for noise filtering \citep{yamada_numerical_2025, yamada_unsupervised_2026}. 
These applications are enabled not only by learning the parameter \(W^{\mathrm{out}}\), but also by interpreting the learned result through the static transformation \(\mathcal{P}\). This suggests that formalizing the static transformation used for interpretation is as important as the parameter learning itself.

The transformation \(\mathcal{S}\), which was introduced to eliminate the residual term \(\Pi_\perp(W^{\mathrm{dyn}})\) appearing in the orthogonal decomposition of the cost, plays a role parallel to that of \(\mathcal{P}\) for \(W^{\mathrm{out}}\).
The parameter \(W^{\mathrm{dyn}}\) obtained by minimizing the reduced cost \(\tilde{\ell}^{\mathrm{sSL}}\) preserves the information needed to recover the supervised readout through \(\mathcal{Q}\). 
That is, \(\mathcal{Q}(W^{\mathrm{dyn}})\) gives the corresponding solution of the supervised reconstruction problem \(W^{\mathrm{out}}r_t \approx s_t\). However, the learned parameter \(W^{\mathrm{dyn}}\) itself does not, in general, define the appropriate learned dynamics via
\begin{alignat}{1}
    \hat{f}^{\mathrm{sSL}}: r \mapsto \sigma\!\left[W^{\mathrm{dyn}}r\right].
\end{alignat}
This is because the reduced cost \(\tilde{\ell}^{\mathrm{sSL}}\) ignores the orthogonal residual component $\Pi_\perp(W^{\mathrm{dyn}})$ that is irrelevant to online learning but necessary for interpreting the result as full reservoir dynamics.
Therefore, the learned parameter should be interpreted through the static transformation \(\mathcal{S}\). 
The corresponding self-supervised learned dynamics is defined by
\begin{alignat}{1}
    \hat{f}^{\mathrm{sSL}}: r \mapsto \sigma\!\left[\mathcal{S}(W^{\mathrm{dyn}})r\right].
\end{alignat}
The transformation \(\mathcal{S}\) can be written as the composition
\begin{alignat}{1}
    \mathcal{S}=\mathcal{P}\circ\mathcal{Q},
\end{alignat}
where \(\mathcal{Q}\) recovers the corresponding supervised readout from \(W^{\mathrm{dyn}}\), and \(\mathcal{P}\) maps this readout back to the parameter space of reservoir dynamics. 
Therefore, the dynamics defined by \(\mathcal{S}(W^{\mathrm{dyn}})\) coincides with the supervised interpretation $\hat{f}^{\mathrm{SL}}$ obtained from the recovered readout \(\mathcal{Q}(W^{\mathrm{dyn}})\). 
Thus, applying \(\mathcal{S}\) to the learned parameter restores the proper dynamical-system interpretation and allows the self-supervised learning result to mimic the external input dynamics \(s_t \mapsto s_{t+1}\), in the same sense as the supervised formulation.

\subsection{A variant for physical implementation}
\label{sec:disc-variant}

In the numerical experiments in Section~\ref{sec:numerical}, we used Eq.~\eqref{eq:Ours} as defined, thereby emphasizing its role as perturbation-based learning for the reduced loss $\tilde{\ell}^{\mathrm{sSL}}$.
The rule, however, contains the transposed pseudoinverse $((\Win)^+)^\top$ as its injection matrix, a globally defined quantity whose explicit use may be problematic in physical implementations.
This issue can be resolved by the following variant.
Replacing the injection matrix $((\Win)^+)^\top$ in Eq.~\eqref{eq:Ours} with $\Win$ yields the update $-\eta\,\delta\ell\,\Win\tilde{\xi}\,r_t^\top$, where $\delta\ell$ denotes the central loss difference appearing in Eq.~\eqref{eq:Ours}.
Since the full-column-rank assumption gives $((\Win)^+)^\top=\Win((\Win)^\top\Win)^{-1}$, this replacement amounts to driving the injection of the original rule with the correlated perturbation $(\Win)^\top\Win\,\tilde{\xi}$ instead of the white perturbation $\tilde{\xi}$.
Both rules leave the $(\operatorname{Im}\Win)^\perp$ component of $\Wdyn$ unchanged and move only within the same affine subspace; restricted to the coordinates of this subspace, their expected updates are gradient descents on $\tilde{\ell}^{\mathrm{sSL}}$ that differ only by the symmetric positive definite preconditioner $(\Win)^\top\Win$.
Consequently, the two rules share the same stationary point, and they coincide exactly when $\Win$ has orthonormal columns; for ill-conditioned $\Win$, the convergence speed, the stable learning-rate range, and the constants of the perturbation-induced variance differ, while the dimension factor remains $O(n_{\mathrm{in}})$.

Under this variant, the perturbation is injected through the existing input pathway as a virtual input: the perturbation pattern $\Win\tilde{\xi}$ is delivered by the fixed input wiring, and no additional routing is required.
Moreover, the loss evaluation reduces to the fixed quadratic form $\frac{1}{2}\|P_{\mathrm{in}}(e_t\pm\alpha\Win\tilde{\xi})\|^2$ of the error $e_t=\Wdyn r_t-\sigma^{-1}(r_{t+1})$, which can be realized as a dedicated circuit that returns a single scalar, for example in analog hardware; $(\Win)^+$ enters the design of the circuit but is never applied explicitly at run time.
The resulting update of each entry, $\Delta\Wdyn_{ij}\propto-\delta\ell\,(\Win\tilde{\xi})_i\,(r_t)_j$, is a three-factor rule that combines presynaptic activity, a postsynaptic perturbation, and a globally broadcast scalar.
Finally, since the transformation above amounts to changing the inner product on the input subspace, it also suggests a natural generalization in which the self-supervised cost in Eq.~\eqref{eq:sSL-cost} is defined from the outset with a general positive-definite metric rather than the standard Euclidean norm. 
The present formulation corresponds to the Euclidean special case. 
We leave this generalization for future work.

\subsection{Assumptions on the reservoir dynamics}
\label{sec:disc-assumptions}

Our theory rests on two regularity assumptions on the reservoir dynamics.
The first is that $\Win$ has full column rank.
The second is that the activation function $\sigma$ is invertible.

The full-column-rank assumption on $\Win$ ensures that $(\Win)^{\top}\Win$ is positive definite.
Therefore, the projected cost $\tilde{\ell}^{\mathrm{sSL}}$ in Eq.~\eqref{eq:OD} is non-degenerate.
Equivalently, $\Win$ is injective as an embedding of $\mathbb{R}^{n_{\mathrm{in}}}$ into $\mathbb{R}^{n_{\mathrm{r}}}$.
This injectivity is what introduces the redundancy discussed above.
In the standard ESN setting, this assumption is essentially free: the entries of $\Win$ are typically drawn at random with $n_{\mathrm{r}}>n_{\mathrm{in}}$, so $\Win$ has full column rank almost surely.

The invertibility of $\sigma$ plays a different role.
It allows the self-supervised learning of $\Wdyn$ to be formulated as a linear regression problem.
It also supports the separation of the online and offline contributions through an orthogonal decomposition induced by a linear projection.
In this sense, the invertibility of $\sigma$ keeps both the learning problem and its decomposition accessible to linear arguments.
The nonlinearity of $\sigma$ is what makes the reservoir a useful information-processing system.
However, its invertibility is also important for the theory of the standard ESN tasks considered here.
Since $\sigma$ acts on each neuron independently, its invertibility is equivalent to the strict monotonicity of each scalar activation function.
This is a genuine restriction: in particular, it excludes ReLU, whose non-injectivity on the negative half-line breaks the invertibility required in our derivation.
Nevertheless, the restriction is mild in practice.
Standard alternatives such as leaky ReLU, ELU, softplus, and tanh are strictly monotone, and hence invertible on their ranges.
Our previous study numerically evaluated the degeneracy of self-supervised learning under a non-invertible activation function \citep{yamada_unsupervised_2026}.
The observed degradation is consistent with the role of invertibility in the present theory.

\subsection{Limitations and future directions}
\label{sec:disc-limitations}

The present study may provide insight into how the low-dimensional structure of a task can be exploited in the design of learning rules.
A limitation of this study, however, is that this idea is demonstrated only in the specific setting of self-supervised learning in ESNs.
An important direction for future work is therefore to examine whether similar redundancy structures can be identified and exploited in more general models and tasks.
Such investigations may clarify how task-specific low-dimensional structures can be incorporated into perturbation-based learning rules beyond ESNs.
A second direction, discussed in Section~\ref{sec:disc-variant}, is to define the self-supervised cost with a general positive-definite metric from the outset, of which the present Euclidean formulation is a special case.
\section{Conclusion}
\label{sec:conclusion}
In this study, we demonstrated that online self-supervised learning in echo state networks admits an orthogonal decomposition of its cost function. 
This decomposition separates the objective into two parts: an input-driven component that must be learned online, and a residual component that can be removed offline by a static transformation determined solely by the fixed reservoir parameters. 
Our findings highlight that the dimensions of the online error representation and perturbation source are governed by the input dimension rather than the reservoir dimension. 
As a result, scalar-feedback perturbation learning becomes scalable to large reservoirs without introducing a reservoir-dimensional factor into the perturbation-induced variance. 
Accordingly, we propose a design principle for online learning rules: when a task has a known low-dimensional structure induced by fixed system parameters, this structure should be actively exploited to restrict the online representation and routing of error signals, as well as the perturbation source, to the dynamically necessary component of the objective.
This perspective reframes self-supervised reformulation. 
It becomes not merely a means of constructing internal targets, but a way of identifying which part of a learning objective genuinely requires online resources. 
These results support a broader view of reservoir computing, in which the fixed structure of the reservoir is treated as exploitable prior knowledge rather than as a constraint to be circumvented. 
More broadly, they point toward local, hardware-compatible learning rules for resource-limited neuromorphic systems, and toward an understanding of how biological systems might leverage their own fixed structure to learn efficiently under scalar, globally broadcast feedback.
\appendix
\section{Proof of the orthogonal decomposition}
\label{app:OD}
\begin{proof}
Let
\begin{equation}
    e_t := \Wdyn r_t-\sigma^{-1}(r_{t+1}),
    \qquad
    P_\mathrm{in}:=\Win(\Win)^+ .
\end{equation}
Since $P_\mathrm{in}$ is the orthogonal projection onto $\operatorname{Im}(\Win)$, it satisfies $P_\mathrm{in}^2=P_\mathrm{in}$ and $P_\mathrm{in}^\top=P_\mathrm{in}$.
Therefore, by the Pythagorean decomposition with respect to
$\mathbb{R}^{n_{\mathrm{r}}}=\operatorname{Im}(\Win)\oplus \operatorname{Im}(\Win)^\perp$, we have
\begin{alignat}{1}
    2\ell^{\mathrm{sSL}}(\Wdyn;r_t,r_{t+1})
    &= \|e_t\|^2 \\
    &= \|P_\mathrm{in}e_t\|^2+\|(I-P_\mathrm{in})e_t\|^2 .
\end{alignat}
The first term is written as
\begin{alignat}{1}
    \|P_\mathrm{in}e_t\|^2
    &= \|\Win(\Win)^+e_t\|^2 \\
    &= \left\|(\Win)^+e_t\right\|^2_{(\Win)^\top\Win} \\
    &= 2\tilde{\ell}^{\mathrm{sSL}}(\Wdyn;r_t,r_{t+1}).
\end{alignat}
For the second term, using $\sigma^{-1}(r_{t+1})=\Win s_t+\Wr r_t$ and $(I-P_\mathrm{in})\Win=0$, we obtain
\begin{alignat}{1}
    \|(I-P_\mathrm{in})e_t\|^2
    &=
    \left\|(I-P_\mathrm{in})(\Wdyn r_t-\Win s_t-\Wr r_t)\right\|^2 \\
    &=
    \left\|(I-P_\mathrm{in})(\Wdyn-\Wr)r_t\right\|^2 \\
    &=
    \left\|\Pi_\perp(\Wdyn)r_t\right\|^2 \\
    &=
    \left\|\Pi_\perp(\Wdyn)\right\|^2_{I,r_t r_t^\top}.
\end{alignat}
Combining the two terms gives
\begin{alignat}{1}
  \ell^{\mathrm{sSL}}(\Wdyn;r_t,r_{t+1})
  =
  \tilde{\ell}^{\mathrm{sSL}}(\Wdyn;r_t,r_{t+1})
  +
  \frac{1}{2}\left\|\Pi_\perp(\Wdyn)\right\|^2_{I,r_t r_t^\top}.
\end{alignat}
This proves Eq.~(\ref{eq:OD}).
\end{proof}

\section{Relations among Learning Rules Induced by the Orthogonal Loss Decomposition}
\label{app:perturbation-relations}
To focus on the essential structure, we introduce abstract notation. Denote the error vector by
\begin{alignat}{1}
    \label{eq:error-vector}
    e=e(\Wdyn):=\Wdyn r_t-\sigma^{-1}(r_{t+1}).
\end{alignat}
The self-supervised learning loss can then be written as $\ell^{\mathrm{sSL}}=\|e\|^2/2$. The particular error vector defined in Eq.~\eqref{eq:error-vector} has the following useful property for any orthogonal projector $P$:
\begin{alignat}{1}
    \label{eq:error-property}
    \frac{\partial}{\partial\Wdyn}\frac{1}{2}\|Pe\|^2=P\frac{\partial}{\partial\Wdyn}\frac{1}{2}\|e\|^2.
\end{alignat}
This identity does not generally hold for an arbitrary matrix $P$ or an arbitrary error vector $e(\Wdyn)$.

For the orthogonal projector $P_{\mathrm{in}}=\Win(\Win)^+$, the projected loss $\|P_{\mathrm{in}}e\|^2/2$ coincides with $\tilde{\ell}^{\mathrm{sSL}}$ in Eq.~\eqref{eq:OD} (see also \ref{app:OD}). We therefore obtain the following relationship between $\mathrm{SGD}(\ell^{\mathrm{sSL}})$ and $\mathrm{SGD}(\tilde{\ell}^{\mathrm{sSL}})$:
\begin{alignat}{1}
    \label{eq:rel-SGD}
    \mathrm{SGD}\left(\frac{1}{2}\|P_{\mathrm{in}}e\|^2\right)=P_{\mathrm{in}}\mathrm{SGD}\left(\frac{1}{2}\|e\|^2\right).
\end{alignat}
Thus, owing to the compatibility between projection and differentiation expressed in Eq.~\eqref{eq:error-property}, $\mathrm{SGD}(\tilde{\ell}^{\mathrm{sSL}})$ can be interpreted as the $P_{\mathrm{in}}$-projection of $\mathrm{SGD}(\ell^{\mathrm{sSL}})$.

Next, we characterize the corresponding relationships for the WP and NP update rules. We assume that the perturbation scale $\alpha$ is sufficiently small for the perturbation-based updates to be treated as approximately unbiased estimators of the corresponding SGD updates. Since WP and NP approximate SGD, they inherit Eq.~\eqref{eq:rel-SGD} in expectation:
\begin{alignat}{1}
    \label{eq:rel-WP-avg}
    \left\langle\mathrm{WP}\left(\frac{1}{2}\|P_{\mathrm{in}}e\|^2\right)\right\rangle_{\Xi}&=\left\langle P_{\mathrm{in}}\mathrm{WP}\left(\frac{1}{2}\|e\|^2\right)\right\rangle_{\Xi},\\
    \label{eq:rel-NP-avg}
    \left\langle\mathrm{NP}\left(\frac{1}{2}\|P_{\mathrm{in}}e\|^2\right)\right\rangle_{\xi}&=\left\langle P_{\mathrm{in}}\mathrm{NP}\left(\frac{1}{2}\|e\|^2\right)\right\rangle_{\xi}.
\end{alignat}
These mean-level identities might suggest corresponding pointwise identities. However, the pointwise updates do not coincide. Defining $Q_{\mathrm{in}}=I-P_{\mathrm{in}}$, their pointwise decompositions are
\begin{alignat}{1}
    \label{eq:rel-WP-point}
    \mathrm{WP}\left(\frac{1}{2}\|P_{\mathrm{in}}e\|^2\right)&=P_{\mathrm{in}}\mathrm{WP}\left(\frac{1}{2}\|P_{\mathrm{in}}e\|^2\right)+\underbrace{Q_{\mathrm{in}}\mathrm{WP}\left(\frac{1}{2}\|P_{\mathrm{in}}e\|^2\right)}_{\text{zero-mean term}},\\
    P_{\mathrm{in}}\mathrm{WP}\left(\frac{1}{2}\|e\|^2\right)&=P_{\mathrm{in}}\mathrm{WP}\left(\frac{1}{2}\|P_{\mathrm{in}}e\|^2\right)+\underbrace{P_{\mathrm{in}}\mathrm{WP}\left(\frac{1}{2}\|Q_{\mathrm{in}}e\|^2\right)}_{\text{zero-mean term}},\\
    \label{eq:rel-NP-point}
    \mathrm{NP}\left(\frac{1}{2}\|P_{\mathrm{in}}e\|^2\right)&=P_{\mathrm{in}}\mathrm{NP}\left(\frac{1}{2}\|P_{\mathrm{in}}e\|^2\right)+\underbrace{Q_{\mathrm{in}}\mathrm{NP}\left(\frac{1}{2}\|P_{\mathrm{in}}e\|^2\right)}_{\text{zero-mean term}},\\
    P_{\mathrm{in}}\mathrm{NP}\left(\frac{1}{2}\|e\|^2\right)&=P_{\mathrm{in}}\mathrm{NP}\left(\frac{1}{2}\|P_{\mathrm{in}}e\|^2\right)+\underbrace{P_{\mathrm{in}}\mathrm{NP}\left(\frac{1}{2}\|Q_{\mathrm{in}}e\|^2\right)}_{\text{zero-mean term}}.
\end{alignat}
Although these additional zero-mean terms leave the expected updates unchanged, they do not generally provide any systematic benefit to the learning rules. For the present purpose, it therefore suffices to retain only the common components $P_{\mathrm{in}}\mathrm{WP}(\|P_{\mathrm{in}}e\|^2/2)$ and $P_{\mathrm{in}}\mathrm{NP}(\|P_{\mathrm{in}}e\|^2/2)$, rather than separately considering $\mathrm{WP}(\|P_{\mathrm{in}}e\|^2/2)$, $P_{\mathrm{in}}\mathrm{WP}(\|e\|^2/2)$, $\mathrm{NP}(\|P_{\mathrm{in}}e\|^2/2)$, and $P_{\mathrm{in}}\mathrm{NP}(\|e\|^2/2)$.

These common components are related to the proposed rule as follows. We define the following canonical-dual perturbation-learning rules:
\begin{alignat}{1}
    \mathrm{cWP}(\tilde{\ell}^{\mathrm{sSL}};\Xi)&:=-\eta\frac{\tilde{\ell}^{\mathrm{sSL}}(\Wdyn+\alpha\Xi;r_t,r_{t+1})-\tilde{\ell}^{\mathrm{sSL}}(\Wdyn-\alpha\Xi;r_t,r_{t+1})}{2\alpha}\operatorname{vec}^{-1}\left(\operatorname{Cov}(\operatorname{vec}(\Xi))^+\operatorname{vec}(\Xi)\right),\\
    \mathrm{cNP}(\tilde{\ell}^{\mathrm{sSL}};\zeta)&:=-\eta\frac{\tilde{\ell}^{\mathrm{sSL}}(y_t+\alpha\zeta;r_{t+1})-\tilde{\ell}^{\mathrm{sSL}}(y_t-\alpha\zeta;r_{t+1})}{2\alpha}\operatorname{Cov}(\zeta)^+\zeta r_t^\top,
\end{alignat}
where $\operatorname{vec}:\mathbb{R}^{n_{\mathrm{r}}\times n_{\mathrm{r}}}\to\mathbb{R}^{n_{\mathrm{r}}^2}$ denotes a fixed vectorization map, and $\operatorname{vec}^{-1}$ denotes its inverse.

Canonical-dual perturbation learning provides a common representation of $P_{\mathrm{in}}\mathrm{WP}(\|P_{\mathrm{in}}e\|^2/2)$, $P_{\mathrm{in}}\mathrm{NP}(\|P_{\mathrm{in}}e\|^2/2)$, and the proposed rule:
\begin{alignat}{1}
    P_{\mathrm{in}}\mathrm{WP}\left(\frac{1}{2}\|P_{\mathrm{in}}e\|^2\right)&=\mathrm{cWP}(\tilde{\ell}^{\mathrm{sSL}};P_{\mathrm{in}}\Xi),\\
    P_{\mathrm{in}}\mathrm{NP}\left(\frac{1}{2}\|P_{\mathrm{in}}e\|^2\right)&=\mathrm{cNP}(\tilde{\ell}^{\mathrm{sSL}};P_{\mathrm{in}}\xi),\\
    \tilde{\Delta}^{\mathrm{Ours}}\Wdyn&=\mathrm{cNP}(\tilde{\ell}^{\mathrm{sSL}};\Win\tilde{\xi}).
\end{alignat}
Thus, both $P_{\mathrm{in}}\mathrm{NP}(\|P_{\mathrm{in}}e\|^2/2)$ and the proposed rule are canonical-dual NP rules, but they employ different perturbations, namely $P_{\mathrm{in}}\xi$ and $\Win\tilde{\xi}$.

Assume that $\xi\sim\mathcal{N}(0,I_{n_{\mathrm{r}}})$ and $\tilde{\xi}\sim\mathcal{N}(0,I_{n_{\mathrm{in}}})$. The random vectors $P_{\mathrm{in}}\xi$ and $\Win\tilde{\xi}$ are then identically distributed if and only if their covariance matrices coincide:
\begin{alignat}{1}
    \operatorname{Cov}(P_{\mathrm{in}}\xi)=\operatorname{Cov}(\Win\tilde{\xi}).
\end{alignat}
Because $\operatorname{Cov}(P_{\mathrm{in}}\xi)=P_{\mathrm{in}}$ and $\operatorname{Cov}(\Win\tilde{\xi})=\Win(\Win)^\top$, this condition is equivalent, under the assumed full column rank of $\Win$, to
\begin{alignat}{1}
    (\Win)^\top\Win=I_{n_{\mathrm{in}}}.
\end{alignat}
Therefore, $P_{\mathrm{in}}\mathrm{NP}(\|P_{\mathrm{in}}e\|^2/2)$ and the proposed rule are identically distributed when $\xi$ and $\tilde{\xi}$ are standard Gaussian random vectors and $\Win$ has orthonormal columns.

In the setting considered here, $\dim\xi=n_{\mathrm{r}}>n_{\mathrm{in}}=\dim\tilde{\xi}$. The proposed update rule in Eq.~\eqref{eq:Ours} can therefore be regarded as canonical-dual perturbation learning driven by the lower-dimensional source of randomness $\tilde{\xi}$.

The term ``canonical dual'' is motivated by frame theory \citep{casazza_finite_2013}, which provides a principled perspective on choosing perturbations according to the structure of the problem. Frame theory also places cWP and cNP within a common framework. A fuller frame-theoretic treatment is beyond the scope of the present study.

For the present purpose, the relationships above justify focusing on the proposed rule rather than treating $\mathrm{WP}(\|P_{\mathrm{in}}e\|^2/2)$, $P_{\mathrm{in}}\mathrm{WP}(\|e\|^2/2)$, $P_{\mathrm{in}}\mathrm{WP}(\|P_{\mathrm{in}}e\|^2/2)$, $\mathrm{NP}(\|P_{\mathrm{in}}e\|^2/2)$, $P_{\mathrm{in}}\mathrm{NP}(\|e\|^2/2)$, and $P_{\mathrm{in}}\mathrm{NP}(\|P_{\mathrm{in}}e\|^2/2)$ as separate methods. All of these rules are constructed using the orthogonal decomposition of the original loss $\ell^{\mathrm{sSL}}$ introduced in Eq.~\eqref{eq:OD} and should therefore be regarded as derived variants of the proposed framework rather than as independent baselines. Accordingly, the numerical experiments compare the proposed method only with the vanilla rules $\mathrm{SGD}(\ell^{\mathrm{sSL}})$, $\mathrm{WP}(\ell^{\mathrm{sSL}})$, and $\mathrm{NP}(\ell^{\mathrm{sSL}})$.

\section{Detailed simulation procedures}
\label{app:simulation}

All simulations were implemented in Python (NumPy, double precision).
Random numbers were drawn from NumPy's pseudo-random generators with explicitly specified seeds, so that every experiment is exactly reproducible.
Throughout the simulations, we used a common generation method for the ESN recurrent matrix $\Wr$, a common activation function $\sigma$, and a common ESN input $s_t$.
Specifically, the entries of $\Wr$ are drawn i.i.d.\ from $\mathcal{N}(0,1)$, and $\Wr$ is then rescaled so that its spectral radius equals $\rho = 0.9$.
For the activation function $\sigma$, we used the element-wise hyperbolic tangent $\tanh$.
For the input, we used a bank of $n_{\mathrm{in}}/2$ sinusoids with distinct frequencies,
\begin{equation}
    s_t = \frac{1}{\sqrt{n_{\mathrm{in}}/2}}
    \left(\cos(a_1 t), \sin(a_1 t), \ldots,
          \cos(a_{n_{\mathrm{in}}/2} t),
          \sin(a_{n_{\mathrm{in}}/2} t)\right)^\top,
    \qquad a_k = k/10,
    \label{eq:input-bank}
\end{equation}
so that the input covariance $\langle s_t s_t^\top \rangle_t$ has full rank $n_{\mathrm{in}}$ and the input norm satisfies $\|s_t\| = 1$ independently of $n_{\mathrm{in}}$.
Unless otherwise noted, we set $n_{\mathrm{in}} = 2$, for which Eq.~(\ref{eq:input-bank}) reduces to $s_t = (\cos(t/10), \sin(t/10))^\top$.

\subsection{Numerical setup for Figure~\ref{fig:orbit}}
\label{app:orbits}
Figure~\ref{fig:orbit} illustrates the orbit of the proposed learning rule defined in Eq.~(\ref{eq:Ours}).
For visualization, we used a reservoir of size $n_\mathrm{r} = 3$ with the input dimension $n_{\mathrm{in}} = 2$, so that the learned parameter $\Wdyn \in \R^{3 \times 3}$ and the first column of $\Wdyn$ can be plotted in $\R^3$.
We used the simple input matrix
\begin{equation}
    \Win =
    \begin{pmatrix}
        1 & 0 \\
        0 & 1 \\
        1 & 1
    \end{pmatrix},
\end{equation}
for which $\mathrm{Im}(\Win)$ and $\mathrm{Im}(\Win)^\perp$ are a known two-dimensional plane and one-dimensional line in $\R^3$, respectively.
Given $(\Win, \Wr, \sigma)$, we generated the reservoir-state sequence $r_t$ by Eq.~(\ref{eq:ESN}) with initial value $r_0 = 0$.
Using $10^5$ steps of this sequence, we applied the proposed learning algorithm [Eq.~(\ref{eq:Ours})] from the two initial values $\Wdyn_0 = O$ and $\Wdyn_0 = \mathcal{S}(O)$, with learning rate $\eta = 10^{-4}$ and perturbation scale $\alpha = 1$.
We then plotted the resulting orbits of $\Wdyn$ together with the subspaces $\mathrm{Im}(\Win)$ and $\mathrm{Im}(\Win)^\perp$.
The optimal solution was computed from the batch solution in Eq.~(\ref{eq:Wdyn*}), and the projected solution was obtained by applying the projection $\Win(\Win)^+$ to the optimal solution.

\subsection{Numerical setup for Figure~\ref{fig:eval}}
\label{app:eval}
Figure~\ref{fig:eval} consists of four panels, A, B, C, and D.

\paragraph{Figure~\ref{fig:eval}A}
The purpose of Figure~\ref{fig:eval}A is to evaluate the signal-to-noise ratio (SNR) of the sample-wise gradient estimates for SGD, WP, NP, and Ours. For a fixed realization $\omega=(\Win,\Wr,\Wdyn)$, let $\widehat{G}_{m,t}(\zeta_m;\omega)$ denote the gradient estimate produced by method $m$ from the state pair $(r_t,r_{t+1})$, where $\zeta_m$ denotes its perturbation. The corresponding reference gradient is
\begin{alignat*}{1}
    \overline{G}_m(\omega)
    =
    \begin{cases}
        \displaystyle \frac{1}{T}\sum_{t=1}^{T} e_t r_t^\top,
        & m\in\{\mathrm{SGD},\mathrm{WP},\mathrm{NP}\},\\[3mm]
        \displaystyle \frac{1}{T}\sum_{t=1}^{T}P_{\mathrm{in}}e_t r_t^\top,
        & m=\mathrm{Ours},
    \end{cases}
\end{alignat*}
where $e_t=W^{\mathrm{dyn}}r_t-\sigma^{-1}(r_{t+1})$ and $P_{\mathrm{in}}=W^{\mathrm{in}}(W^{\mathrm{in}})^+$. 
We define the conditional SNR for this realization as
\begin{alignat}{1}
    \operatorname{SNR}_m(\omega)
    =
    \frac{\|\overline{G}_m(\omega)\|_{\mathrm{F}}^2}
    {\left\langle
    \left[
        \left\|
            \widehat{G}_{m,t}(\zeta_m;\omega)
            -\overline{G}_m(\omega)
        \right\|_{\mathrm{F}}^2
        \,\middle|\,\omega
    \right]\right\rangle_{t,\zeta_m}}.
    \label{eq:SNR}
\end{alignat}
Here, $t$ is sampled uniformly from the retained state pairs, and the expectation is additionally taken over the perturbation for WP, NP, and Ours. For SGD, no perturbation is drawn. 
Thus, the denominator includes both variability across state pairs and, where applicable, the additional variability induced by perturbation. The matrices $\Win$, $\Wr$, and $\Wdyn$, as well as the reservoir trajectory, are held fixed within Eq.~(\ref{eq:SNR}); variability across their realizations is summarized only after computing the SNR separately for each realization. All norms are Frobenius norms in the full parameter space.

For each reservoir size $n_\mathrm{r}=2^k$ ($k=1,\ldots,10$), with $n_{\mathrm{in}}=2$, we generated $\Wr$ as described above and independently generated $\Win$ and $\Wdyn$ with entries drawn i.i.d.\ from $\mathcal{N}(0,1)$.
After discarding the first $10^2$ transient steps, the reference gradient was computed from all $T=10^4$ retained state pairs. The expectation in the denominator of Eq.~(\ref{eq:SNR}) was estimated from 50 time indices sampled uniformly with replacement. A fresh independent standard Gaussian perturbation with scale $\alpha=1$ was drawn for each sample of WP, NP, and Ours, whereas no perturbation was used for SGD. We repeated this procedure for 50 realizations of $(W^{\mathrm{in}},W^{\mathrm{r}},W^{\mathrm{dyn}})$ and report the median and interquartile range of the resulting SNRs.

\paragraph{Figure~\ref{fig:eval}B}
The purpose of Figure~\ref{fig:eval}B is to evaluate the dependence of the SNR on the input dimension. We fixed $n_\mathrm{r}=128$ and varied the input dimension as $n_{\mathrm{in}}=2^k$ ($k=1,\ldots,7$), with the largest value corresponding to $n_{\mathrm{in}}=n_\mathrm{r}$. To avoid numerical instability caused by an ill-conditioned pseudoinverse, $W^{\mathrm{in}}$ was generated with orthonormal columns by applying QR decomposition to a random Gaussian matrix. All other procedures were identical to those used for Figure~\ref{fig:eval}A: the reference gradient was computed from $T=10^4$ retained state pairs, the denominator was estimated from 50 sampled gradient estimates, and the median and interquartile range were calculated across 50 realizations.

\paragraph{Figure~\ref{fig:eval}C}
The purpose of Figure~\ref{fig:eval}C is to numerically evaluate the optimal learning rate for a finite number of iterations.
We fixed $n_\mathrm{r} = 100$ and, for each of $50$ realizations of $(\Win, \Wr)$, generated a reservoir-state sequence of $10^2 + 10^4 + 10^4$ steps by driving the ESN with the input $s_t$ (Eq.~\ref{eq:ESN}).
The first $10^2$ steps were discarded as a transient, the next $10^4$ steps were used for training, and the last $10^4$ steps for evaluation.
Starting from $\Wdyn_0 = O$, we trained each algorithm (SGD, WP, NP, and Ours) for $10^4$ steps at each of $400$ learning rates $\eta$ spaced logarithmically over $[10^{-7}, 10^1]$.
For each run, we computed the sample mean of the projected loss $\tilde{\ell}^{\mathrm{sSL}}$ over the evaluation steps, normalized it so that its value at the initial point $\Wdyn_0 = O$ equals $1$, and plotted the median and interquartile range across realizations against $\eta$.
The learning rate minimizing the median normalized loss was recorded for each algorithm ($\eta = 3.56 \times 10^{-3}$ for SGD, $1.61 \times 10^{-5}$ for WP, $1.22 \times 10^{-4}$ for NP, and $1.98 \times 10^{-1}$ for Ours) and used in the simulations for Figure~\ref{fig:eval}D.

\paragraph{Figure~\ref{fig:eval}D}
Figure~\ref{fig:eval}D shows the learning curves at the empirically optimal learning rates obtained from the simulations for Figure~\ref{fig:eval}C.
We again fixed $n_\mathrm{r} = 100$ and, for each of $50$ realizations of $(\Win, \Wr)$, generated a reservoir-state sequence of $10^2 + 10^5 + 10^3$ steps by driving the ESN with the input $s_t$ (Eq.~\ref{eq:ESN}).
The first $10^2$ steps were discarded as a transient, the next $10^5$ steps were used for training, and the last $10^3$ steps for evaluation.
Starting from $\Wdyn_0 = O$, we trained each algorithm (SGD, WP, NP, and Ours) for $10^5$ steps at the learning rate obtained for that algorithm in Figure~\ref{fig:eval}C.
During training, we computed the sample mean of the projected loss $\tilde{\ell}^{\mathrm{sSL}}$ over the evaluation steps, normalized it so that its value at the initial point $\Wdyn_0 = O$ equals $1$, and plotted the median and interquartile range across realizations against the number of iterations.

\section*{Acknowledgments}
This work was partially supported by JSPS KAKENHI Grant Numbers JP22K18419, JP24K15161, JP25H00451, and JP25K24744, JSPS Research Fellowship for Young Scientists Grant Number JP25KJ0990, JST Moonshot R\&D Grant Number JPMJMS2021, JST ACT-X Grant Number JPMJAX24CT and a project, JPNP14004, commissioned by the New Energy and Industrial Technology Development Organization (NEDO).

\bibliographystyle{elsarticle-harv}
\bibliography{PL-in-ESN, ref}

@article{hara_learning_2022,
	title = {Learning {Dynamics} by {Reservoir} {Computing} ({In} {Memory} of {Prof}. {Pavol} {Brunovský})},
	issn = {1572-9222},
	url = {https://doi.org/10.1007/s10884-022-10159-w},
	doi = {10.1007/s10884-022-10159-w},
	language = {en},
	urldate = {2023-02-24},
	journal = {Journal of Dynamics and Differential Equations},
	author = {Hara, Masato and Kokubu, Hiroshi},
	month = apr,
	year = {2022},
}

@article{pathak_model-free_2018,
	title = {Model-{Free} {Prediction} of {Large} {Spatiotemporally} {Chaotic} {Systems} from {Data}: {A} {Reservoir} {Computing} {Approach}},
	volume = {120},
	shorttitle = {Model-{Free} {Prediction} of {Large} {Spatiotemporally} {Chaotic} {Systems} from {Data}},
	url = {https://link.aps.org/doi/10.1103/PhysRevLett.120.024102},
	doi = {10.1103/PhysRevLett.120.024102},
	number = {2},
	urldate = {2023-02-24},
	journal = {Physical Review Letters},
	publisher = {American Physical Society},
	author = {Pathak, Jaideep and Hunt, Brian and Girvan, Michelle and Lu, Zhixin and Ott, Edward},
	month = jan,
	year = {2018},
	note = {Number: 2},
	pages = {024102},
}

@article{lu_reservoir_2017,
	title = {Reservoir observers: {Model}-free inference of unmeasured variables in chaotic systems},
	volume = {27},
	issn = {1054-1500},
	shorttitle = {Reservoir observers},
	url = {https://aip.scitation.org/doi/full/10.1063/1.4979665},
	doi = {10.1063/1.4979665},
	number = {4},
	urldate = {2023-02-24},
	journal = {Chaos: An Interdisciplinary Journal of Nonlinear Science},
	publisher = {American Institute of Physics},
	author = {Lu, Zhixin and Pathak, Jaideep and Hunt, Brian and Girvan, Michelle and Brockett, Roger and Ott, Edward},
	month = apr,
	year = {2017},
	note = {Number: 4},
	pages = {041102},
}

@article{maass_real-time_2002,
	title = {Real-{Time} {Computing} {Without} {Stable} {States}: {A} {New} {Framework} for {Neural} {Computation} {Based} on {Perturbations}},
	volume = {14},
	issn = {0899-7667, 1530-888X},
	shorttitle = {Real-{Time} {Computing} {Without} {Stable} {States}},
	url = {https://direct.mit.edu/neco/article/14/11/2531-2560/6650},
	doi = {10.1162/089976602760407955},
	language = {en},
	number = {11},
	urldate = {2024-02-21},
	journal = {Neural Computation},
	author = {Maass, Wolfgang and Natschläger, Thomas and Markram, Henry},
	month = nov,
	year = {2002},
	pages = {2531--2560},
}

@article{tanaka_recent_2019,
	title = {Recent advances in physical reservoir computing: {A} review},
	volume = {115},
	issn = {0893-6080},
	shorttitle = {Recent advances in physical reservoir computing},
	url = {https://www.sciencedirect.com/science/article/pii/S0893608019300784},
	doi = {10.1016/j.neunet.2019.03.005},
	urldate = {2025-03-28},
	journal = {Neural Networks},
	author = {Tanaka, Gouhei and Yamane, Toshiyuki and Héroux, Jean Benoit and Nakane, Ryosho and Kanazawa, Naoki and Takeda, Seiji and Numata, Hidetoshi and Nakano, Daiju and Hirose, Akira},
	month = jul,
	year = {2019},
	pages = {100--123},
}

@article{yamada_unsupervised_2026,
	title = {Unsupervised {Learning} in {Echo} {State} {Networks} for {Input} {Reconstruction}},
	volume = {38},
	issn = {0899-7667},
	url = {https://doi.org/10.1162/NECO.a.38},
	doi = {10.1162/NECO.a.38},
	number = {2},
	urldate = {2026-05-15},
	journal = {Neural Computation},
	author = {Yamada, Taiki and Katori, Yuichi and Fujiwara, Kantaro},
	month = jan,
	year = {2026},
	pages = {198--227},
}

@article{indiveri_memory_2015,
	title = {Memory and {Information} {Processing} in {Neuromorphic} {Systems}},
	volume = {103},
	issn = {1558-2256},
	url = {https://ieeexplore.ieee.org/document/7159144},
	doi = {10.1109/JPROC.2015.2444094},
	number = {8},
	urldate = {2026-07-06},
	journal = {Proceedings of the IEEE},
	author = {Indiveri, Giacomo and Liu, Shih-Chii},
	month = aug,
	year = {2015},
	pages = {1379--1397},
}

@article{parisi_continual_2019,
	title = {Continual lifelong learning with neural networks: {A} review},
	volume = {113},
	issn = {0893-6080},
	shorttitle = {Continual lifelong learning with neural networks},
	url = {https://www.sciencedirect.com/science/article/pii/S0893608019300231},
	doi = {10.1016/j.neunet.2019.01.012},
	urldate = {2026-07-06},
	journal = {Neural Networks},
	author = {Parisi, German I. and Kemker, Ronald and Part, Jose L. and Kanan, Christopher and Wermter, Stefan},
	month = may,
	year = {2019},
	pages = {54--71},
}

@article{mead_neuromorphic_1990,
	title = {Neuromorphic electronic systems},
	volume = {78},
	issn = {1558-2256},
	url = {https://ieeexplore.ieee.org/document/58356},
	doi = {10.1109/5.58356},
	number = {10},
	urldate = {2026-07-06},
	journal = {Proceedings of the IEEE},
	author = {Mead, C.},
	month = oct,
	year = {1990},
	pages = {1629--1636},
}

@article{lukosevicius_reservoir_2009,
	title = {Reservoir computing approaches to recurrent neural network training},
	volume = {3},
	issn = {1574-0137},
	url = {https://www.sciencedirect.com/science/article/pii/S1574013709000173},
	doi = {10.1016/j.cosrev.2009.03.005},
	number = {3},
	urldate = {2026-07-06},
	journal = {Computer Science Review},
	author = {Lukoševičius, Mantas and Jaeger, Herbert},
	month = aug,
	year = {2009},
	pages = {127--149},
}

@inproceedings{jaeger_adaptive_2002,
	title = {Adaptive {Nonlinear} {System} {Identification} with {Echo} {State} {Networks}},
	volume = {15},
	url = {https://proceedings.neurips.cc/paper_files/paper/2002/hash/426f990b332ef8193a61cc90516c1245-Abstract.html},
	urldate = {2026-07-06},
	booktitle = {Advances in {Neural} {Information} {Processing} {Systems}},
	publisher = {MIT Press},
	author = {Jaeger, Herbert},
	year = {2002},
}

@article{dambre_information_2012,
	title = {Information {Processing} {Capacity} of {Dynamical} {Systems}},
	volume = {2},
	copyright = {2012 The Author(s)},
	issn = {2045-2322},
	url = {https://www.nature.com/articles/srep00514},
	doi = {10.1038/srep00514},
	language = {en},
	number = {1},
	urldate = {2026-07-06},
	journal = {Scientific Reports},
	publisher = {Nature Publishing Group},
	author = {Dambre, Joni and Verstraeten, David and Schrauwen, Benjamin and Massar, Serge},
	month = jul,
	year = {2012},
	pages = {514},
}

@inproceedings{lin_-device_2022,
	title = {On-{Device} {Training} {Under} {256KB} {Memory}},
	volume = {35},
	url = {https://proceedings.neurips.cc/paper_files/paper/2022/hash/90c56c77c6df45fc8e556a096b7a2b2e-Abstract-Conference.html},
	urldate = {2026-07-06},
	booktitle = {Advances in {Neural} {Information} {Processing} {Systems}},
	publisher = {Curran Associates, Inc.},
	author = {Lin, Ji and Zhu, Ligeng and Chen, Wei-Ming and Wang, Wei-Chen and Gan, Chuang and Han, Song},
	year = {2022},
	pages = {22941--22954},
}

@article{lillicrap_backpropagation_2020,
	title = {Backpropagation and the brain},
	volume = {21},
	copyright = {2020 Springer Nature Limited},
	issn = {1471-0048},
	url = {https://www.nature.com/articles/s41583-020-0277-3},
	doi = {10.1038/s41583-020-0277-3},
	language = {en},
	number = {6},
	urldate = {2026-07-06},
	journal = {Nature Reviews Neuroscience},
	publisher = {Nature Publishing Group},
	author = {Lillicrap, Timothy P. and Santoro, Adam and Marris, Luke and Akerman, Colin J. and Hinton, Geoffrey},
	month = jun,
	year = {2020},
	pages = {335--346},
}

@article{jaeger_harnessing_2004,
	title = {Harnessing {Nonlinearity}: {Predicting} {Chaotic} {Systems} and {Saving} {Energy} in {Wireless} {Communication}},
	volume = {304},
	shorttitle = {Harnessing {Nonlinearity}},
	url = {https://www.science.org/doi/10.1126/science.1091277},
	doi = {10.1126/science.1091277},
	number = {5667},
	urldate = {2026-07-06},
	journal = {Science},
	publisher = {American Association for the Advancement of Science},
	author = {Jaeger, Herbert and Haas, Harald},
	month = apr,
	year = {2004},
	pages = {78--80},
}

@article{sussillo_generating_2009,
	title = {Generating coherent patterns of activity from chaotic neural networks},
	volume = {63},
	issn = {1097-4199},
	doi = {10.1016/j.neuron.2009.07.018},
	language = {eng},
	number = {4},
	journal = {Neuron},
	author = {Sussillo, David and Abbott, L. F.},
	month = aug,
	year = {2009},
	pages = {544--557},
}

@article{plackett_theorems_1950,
	title = {Some {Theorems} in {Least} {Squares}},
	volume = {37},
	issn = {0006-3444},
	url = {https://www.jstor.org/stable/2332158},
	doi = {10.2307/2332158},
	number = {1/2},
	urldate = {2026-07-06},
	journal = {Biometrika},
	publisher = {[Oxford University Press, Biometrika Trust]},
	author = {Plackett, R. L.},
	year = {1950},
	pages = {149--157},
}

@article{robbins_stochastic_1951,
	title = {A {Stochastic} {Approximation} {Method}},
	volume = {22},
	issn = {0003-4851, 2168-8990},
	url = {https://projecteuclid.org/journals/annals-of-mathematical-statistics/volume-22/issue-3/A-Stochastic-Approximation-Method/10.1214/aoms/1177729586.full},
	doi = {10.1214/aoms/1177729586},
	language = {en},
	number = {3},
	urldate = {2026-07-06},
	journal = {The Annals of Mathematical Statistics},
	publisher = {Institute of Mathematical Statistics},
	author = {Robbins, Herbert and Monro, Sutton},
	month = sep,
	year = {1951},
	pages = {400--407},
}

@inproceedings{bottou_large-scale_2010,
	address = {Heidelberg},
	title = {Large-{Scale} {Machine} {Learning} with {Stochastic} {Gradient} {Descent}},
	isbn = {978-3-7908-2604-3},
	doi = {10.1007/978-3-7908-2604-3_16},
	language = {en},
	booktitle = {Proceedings of {COMPSTAT}'2010},
	publisher = {Physica-Verlag HD},
	author = {Bottou, Léon},
	editor = {Lechevallier, Yves and Saporta, Gilbert},
	year = {2010},
	pages = {177--186},
}

@article{kiefer_stochastic_1952,
	title = {Stochastic {Estimation} of the {Maximum} of a {Regression} {Function}},
	volume = {23},
	issn = {0003-4851, 2168-8990},
	url = {https://projecteuclid.org/journals/annals-of-mathematical-statistics/volume-23/issue-3/Stochastic-Estimation-of-the-Maximum-of-a-Regression-Function/10.1214/aoms/1177729392.full},
	doi = {10.1214/aoms/1177729392},
	language = {en},
	number = {3},
	urldate = {2026-07-06},
	journal = {The Annals of Mathematical Statistics},
	publisher = {Institute of Mathematical Statistics},
	author = {Kiefer, J. and Wolfowitz, J.},
	month = sep,
	year = {1952},
	pages = {462--466},
}

@inproceedings{flower_summed_1992,
	title = {Summed {Weight} {Neuron} {Perturbation}: {An} {O}({N}) {Improvement} {Over} {Weight} {Perturbation}},
	volume = {5},
	shorttitle = {Summed {Weight} {Neuron} {Perturbation}},
	url = {https://proceedings.neurips.cc/paper/1992/hash/996a7fa078cc36c46d02f9af3bef918b-Abstract.html},
	urldate = {2026-07-06},
	booktitle = {Advances in {Neural} {Information} {Processing} {Systems}},
	publisher = {Morgan-Kaufmann},
	author = {Flower, Barry and Jabri, Marwan},
	year = {1992},
}

@article{fiete_gradient_2006,
	title = {Gradient learning in spiking neural networks by dynamic perturbation of conductances},
	volume = {97},
	issn = {0031-9007},
	doi = {10.1103/PhysRevLett.97.048104},
	language = {eng},
	number = {4},
	journal = {Physical Review Letters},
	author = {Fiete, Ila R. and Seung, H. Sebastian},
	month = jul,
	year = {2006},
	pages = {048104},
}

@article{werfel_learning_2005,
	title = {Learning curves for stochastic gradient descent in linear feedforward networks},
	volume = {17},
	issn = {0899-7667},
	doi = {10.1162/089976605774320539},
	language = {eng},
	number = {12},
	journal = {Neural Computation},
	author = {Werfel, Justin and Xie, Xiaohui and Seung, H. Sebastian},
	month = dec,
	year = {2005},
	pages = {2699--2718},
}

@article{yamada_numerical_2025,
	title = {Numerical {Evaluation} of a {Weakly} {Supervised} {Filtering} {Method} {Based} on {Echo} {State} {Networks}},
	volume = {2026},
	doi = {10.5687/sss.2026.109},
	journal = {Proceedings of the ISCIE International Symposium on Stochastic Systems Theory and its Applications},
	author = {Yamada, Taiki and Katori, Yuichi and Fujiwara, Kantaro},
	year = {2025},
	pages = {109--113},
}

@article{zuge_weight_2023,
	title = {Weight versus {Node} {Perturbation} {Learning} in {Temporally} {Extended} {Tasks}: {Weight} {Perturbation} {Often} {Performs} {Similarly} or {Better}},
	volume = {13},
	shorttitle = {Weight versus {Node} {Perturbation} {Learning} in {Temporally} {Extended} {Tasks}},
	url = {https://link.aps.org/doi/10.1103/PhysRevX.13.021006},
	doi = {10.1103/PhysRevX.13.021006},
	number = {2},
	urldate = {2026-07-13},
	journal = {Physical Review X},
	publisher = {American Physical Society},
	author = {Züge, Paul and Klos, Christian and Memmesheimer, Raoul-Martin},
	month = apr,
	year = {2023},
	pages = {021006},
}

@book{casazza_finite_2013,
	address = {Boston},
	series = {Applied and {Numerical} {Harmonic} {Analysis}},
	title = {Finite {Frames}: {Theory} and {Applications}},
	copyright = {https://www.springernature.com/gp/researchers/text-and-data-mining},
	isbn = {978-0-8176-8372-6 978-0-8176-8373-3},
	shorttitle = {Finite {Frames}},
	url = {https://link.springer.com/10.1007/978-0-8176-8373-3},
	doi = {10.1007/978-0-8176-8373-3},
	language = {en},
	urldate = {2026-07-22},
	publisher = {Birkhäuser},
	editor = {Casazza, Peter G. and Kutyniok, Gitta},
	year = {2013},
}

@article{jabri_weight_1991,
	title = {Weight {Perturbation}: {An} {Optimal} {Architecture} and {Learning} {Technique} for {Analog} {VLSI} {Feedforward} and {Recurrent} {Multilayer} {Networks}},
	volume = {3},
	issn = {0899-7667},
	shorttitle = {Weight {Perturbation}},
	url = {https://ieeexplore.ieee.org/document/6796476},
	doi = {10.1162/neco.1991.3.4.546},
	number = {4},
	urldate = {2026-07-23},
	journal = {Neural Computation},
	author = {Jabri, Marwan and Flower, Barry},
	month = feb,
	year = {1991},
	pages = {546--565},
}

@article{jaeger2001echo,
  title={The “echo state” approach to analysing and training recurrent neural networks},
  author={Jaeger, Herbert},
  journal={Bonn, Germany: German National Research Center for Information Technology GMD Technical Report},
  volume={148},
  number={34},
  pages={13},
  year={2001},
}

@book{Gauss1809,
  author    = {Gauss, Carl Friedrich},
  title     = {Theoria motus corporum coelestium in sectionibus conicis solem ambientium},
  publisher = {Frid. Perthes et I. H. Besser},
  address   = {Hamburg},
  year      = {1809}
}

@book{Legendre1805,
  author    = {Legendre, Adrien-Marie},
  title     = {Nouvelles m{\'e}thodes pour la d{\'e}termination des orbites des com{\`e}tes},
  publisher = {F. Didot},
  address   = {Paris},
  year      = {1805}
}

@article{Cauchy1847,
  author  = {Cauchy, Augustin-Louis},
  title   = {M{\'e}thode g{\'e}n{\'e}rale pour la r{\'e}solution des syst{\`e}mes d'{\'e}quations simultan{\'e}es},
  journal = {Comptes rendus hebdomadaires des s{\'e}ances de l'Acad{\'e}mie des sciences},
  volume  = {25},
  pages   = {536--538},
  year    = {1847}
}

@inproceedings{
ren2023scaling,
title={Scaling Forward Gradient With Local Losses},
author={Mengye Ren and Simon Kornblith and Renjie Liao and Geoffrey Hinton},
booktitle={The Eleventh International Conference on Learning Representations },
year={2023},
url={https://openreview.net/forum?id=JxpBP1JM15-}
}

\clearpage
\setcounter{section}{0}
\setcounter{figure}{0}
\setcounter{table}{0}
\setcounter{equation}{0}
\renewcommand{\thesection}{S\arabic{section}}
\renewcommand{\thefigure}{S\arabic{figure}}
\renewcommand{\thetable}{S\arabic{table}}
\renewcommand{\theequation}{S\arabic{equation}}   
\section*{Supplementary Material}
\begin{figure}[H]
    \centering
    \includegraphics[width=\textwidth]{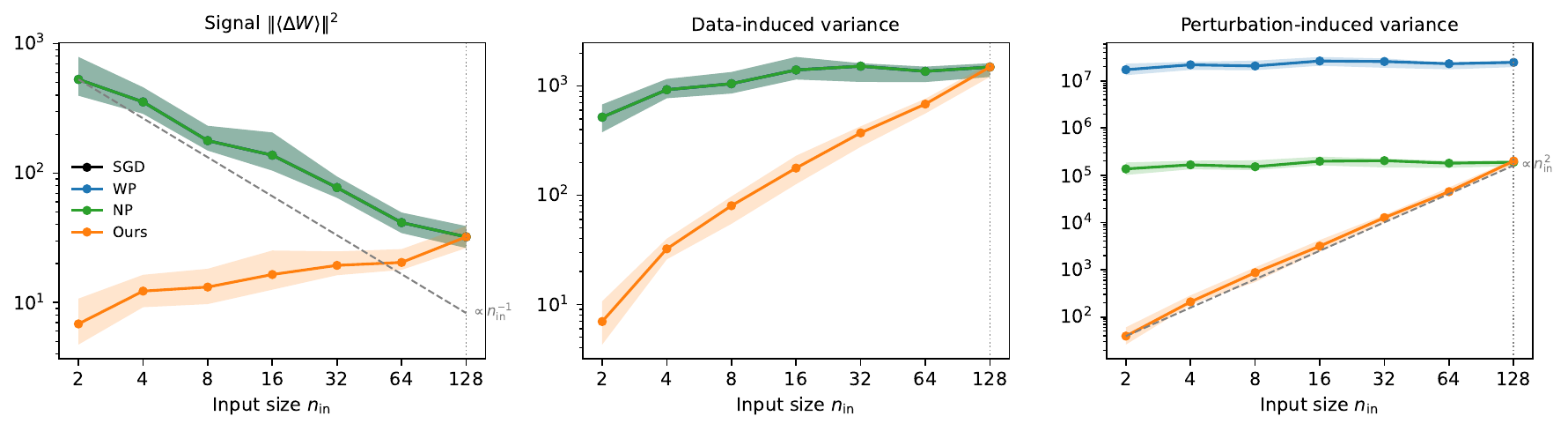}
    \caption{%
Decomposition of the update-rule fluctuations over the input
dimension ($n_\mathrm{r}=128$; $W^{\mathrm{in}}$ has orthonormal columns).
The squared deviation of each stochastic update from its time-averaged
reference is decomposed into data-induced and perturbation-induced components.
(Left) For the full-loss rules SGD, WP, and NP, the signal
$\|\langle\Delta W^{\mathrm{dyn}}\rangle\|^2$ decreases approximately as
$n_{\mathrm{in}}^{-1}$, explaining their common downward SNR trend in
Figure~\ref{fig:eval}B. (Middle) Their data-induced variance depends only weakly on
$n_{\mathrm{in}}$. (Right) The perturbation-induced variance is zero for SGD,
approximately independent of $n_{\mathrm{in}}$ for WP and NP, and increases
with $n_{\mathrm{in}}$ for Ours. At $n_{\mathrm{in}}=n_\mathrm{r}$, the quantities for
Ours coincide with those for vanilla NP under the Gaussian-perturbation
setting, because the projection onto $\operatorname{Im}\Win$ becomes
the identity.}
    \label{fig:supp_decomposition}
\end{figure}
\end{document}